\documentclass[twoside,11pt]{article}

\usepackage{jmlr2e}

\usepackage{microtype}
\usepackage{graphicx}
\usepackage{subfigure}
\usepackage{booktabs} 
\usepackage{amsmath,amsfonts,amssymb}
\usepackage[english]{babel}
\usepackage[squaren]{SIunits}
\usepackage{todonotes}
\usepackage{algorithm}
\usepackage{enumerate}
\usepackage[noend]{algpseudocode}

\jmlrheading{21}{01}{2020}{}{}{F. Tonolini, J. Radford, A. Turpin, D. Faccio and R. Murray-Smith}

\ShortHeadings{Variational Inference for Computational Imaging Inverse Problems}{Tonolini, Radford, Turpin, Faccio and Murray-Smith}
\firstpageno{1}

\begin{document}

\title{Variational Inference for Computational Imaging Inverse Problems}

\author{\name Francesco Tonolini \email 2402432t@student.gla.ac.uk \\
       \addr School of Computing Science, 
       University of Glasgow
       \AND
       \name Jack Radford \email j.radford.1@research.gla.ac.uk \\
       \addr School of Physics and Astronomy,
       University of Glasgow
      \AND
       \name Alex Turpin \email alex.turpin@glasgow.ac.uk \\
       \addr School of Computing Science,
       University of Glasgow
      \AND
       \name Daniele Faccio \email daniele.faccio@glasgow.ac.uk \\
       \addr School of Physics and Astronomy,
       University of Glasgow
       \AND
       \name Roderick Murray-Smith \email roderick.murray-smith@glasgow.ac.uk \\
       \addr School of Computing Science,
       University of Glasgow}

\editor{}

\maketitle

\begin{abstract}
Machine learning methods for computational imaging require uncertainty estimation to be reliable in real settings. While Bayesian models offer a computationally tractable way of recovering uncertainty,
they need large data volumes to be trained, which in imaging applications implicates prohibitively expensive collections with specific imaging instruments. This paper introduces a novel framework to train variational inference for inverse problems exploiting in combination few experimentally collected data, domain expertise and existing image data sets. In such a way, Bayesian machine learning models can solve imaging inverse problems with minimal data collection efforts.
Extensive simulated experiments show the advantages of the proposed framework.
The approach is then applied to two real experimental optics settings: holographic image reconstruction and imaging through highly scattering media. In both settings, state of the art reconstructions are achieved with little collection of training data.

\end{abstract}


\section{Introduction}
\label{intro}

Computational imaging (CI)
is one of the most important yet challenging forms of algorithmic information retrieval, with applications in Medicine, Biology, Astronomy and more.
Bayesian machine learning methods are an attractive route to solve CI inverse problems, as they retain the advantages of learning from empirical data, while improving reliability by inferring uncertainty \citep{AH,PRR}. However,
fitting distributions with Bayesian models requires large sets of training examples \citep{AVI}. This is particularly problematic in CI settings, where measurements are often unique to specific instruments, resulting in the necessity to carry out lengthy and expensive acquisition experiments or simulations to collect training data \citep{NNIP,LIM_DATA1}. Consider, for example, the task of reconstructing three dimensional environments from LIDAR measurements. Applying machine learning to this task requires data, in particular, paired examples of 3D environments and signals recorded with the particular LIDAR system to be employed. This means that, in principle, each LIDAR system being developed for this task requires its own extensive data set of paired examples to be collected, rendering the use of machine learning extremely impractical.

\begin{figure*}[t!]
  \centering
  \includegraphics[width=\linewidth]{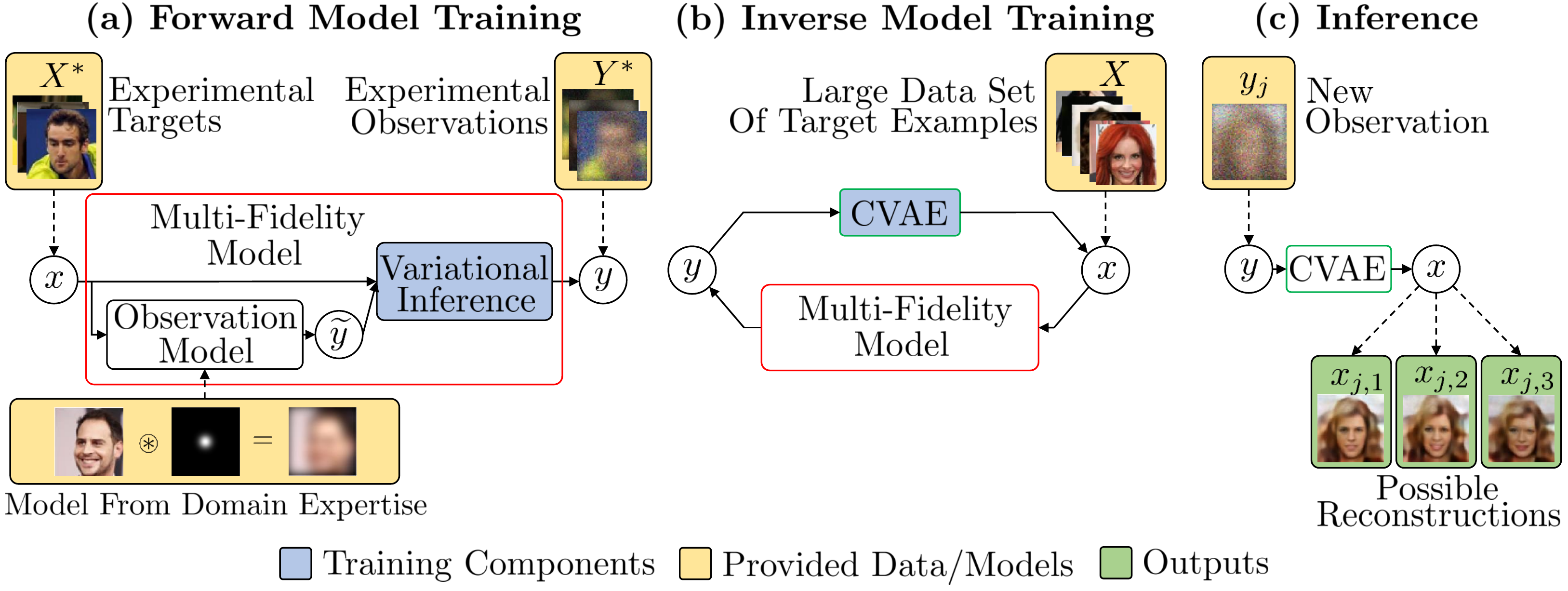}
\caption{Proposed framework for training variational inference with diverse sources of information accessible in imaging settings. (a) Firstly, a multi-fidelity forward model is built to generate experimental observations. A variational model is trained to reproduce experimental observations $Y^*$ from experimental ground truth targets $X^*$, exploiting simulated predictions $\widetilde{y}$
 given by some analytical observation model defined with domain expertise. (b) A CVAE then learns to solve the inverse problem from a large data set of target examples $X$ with a training loop; target examples $x$ are passed through the previously learned multi-fidelity forward model to generate measurements $y$, which are then used as conditions for training the CVAE to generate back the targets $x$. In this way, a large number of ground truth targets can be exploited for learning, without the need for associated experimental measurements. (c) The trained CVAE can then be used to draw different possible solutions $x_{j,i}$ to the inverse problem conditioned on a new observation $y_j$.
}
\label{fig:models_2}
\end{figure*}

This paper introduces a novel framework to train conditional variational models for solving CI inverse problems leveraging in combination (i) a minimal amount of experimentally acquired or numerically simulated ground truth target-observation pairs, (ii) an inexpensive analytical model of the observation process from domain expertise and (iii) a large number of unobserved target examples, which can often be found in existing data sets. In such a way, trained variational inference models benefit from all accessible useful data as well as domain expertise, rather than relying solely on specifically collected training inputs and outputs.
Recalling the LIDAR example given above, the proposed method would allow the joint utilisation of limited collections with the specific LIDAR instrument, a physical model of LIDAR acquisition and any number of available examples of 3D environments to train our machine learning models.
The framework is derived with Bayesian formulations, interpreting the different sources of information available as samples from or approximations to the underlying hidden distributions.

To address the expected scarcity of experimental data, the novel training strategy adopts a variational semi-supervision approach. Similarly to recent works in semi-supervised VAEs, an auxiliary model is employed to map abundantly available target ground-truths to corresponding measurements, which are, in contrast, scarce \citep{SEM_GEN_1,SEM_GEN_2,SEM_GEN_3}. However, the proposed framework introduces two important differences, specifically adapting to CI problems:
\begin{enumerate}[i]
    \item This auxiliary function incorporates a physical observation model designed with domain expertise. In most imaging settings, the Physics of how targets map to corresponding measurements is well understood and described by closed form expressions. These should be used to improve the quality of a reconstruction system.
    \item Instead of training the two models simultaneously, a forward model is trained first and then employed as a sampler in training the reconstruction model.  This choice is made to avoid that synthetic measurements, i.e. predicted by the auxiliary system, contain more information about the targets than those encountered in reality. While this is not a critical problem for most semi-supervised systems, as auxiliary models often predict low-dimensional conditions such as labels, it very much arises in imaging settings, where these conditions are instead measurements that have comparable or even higher dimensionality than the targets. This high dimensionality of the conditions allows a system training the two models jointly to pass rich information through the synthetic measurements in order to maximise training reconstruction likelihood. By training the forward process separately instead, this auxiliary model is induced to maximise fidelity to real measurements alone, essentially providing an emulator. 
\end{enumerate}
Figure \ref{fig:models_2} schematically illustrate the framework and its components.

The proposed framework is first quantitatively evaluated in simulated image recovery experiments, making use of the benchmark data sets CelebA and CIFAR10 \citep{celebA, CIFAR}. In these experiments, different common transformations are applied to the images, including Gaussian blurring, partial occlusion and down-sampling. Image recovery is then performed with variational models. The novel training framework proved significantly advantageous across the range of tested conditions compared with standard training strategies. 
Secondly, the proposed technique is implemented with experimental imaging systems in two different settings; phase-less holographic image reconstruction \citep{2015:ieee:segev,2017:optica:barbastathis} and imaging through highly scattering media with temporally resolved sensors \citep{DOT_BOOK,DOT_NAT}. 
In both settings, state of the art results are obtained with the proposed technique, while requiring minimal experimental collection efforts for training.

\section{Background and Related Work}

Computational imaging (CI) broadly identifies a class of methods in which an image of interest is not directly observed, but rather inferred algorithmically from one or more measurements \citep{CIBOOK1}. Many image recovery tasks fall within this definition, such as de-convolution \citep{DECONV,DECONV_TV}, computed tomography (CT) \citep{CT1} and structured illumination imaging \citep{SP1}. Traditionally, CI tasks are modelled as inverse problems, where a target signal $x \in \mathbb{R}^n$ is measured through a forward model $y = f(x)$, yielding observations $y \in \mathbb{R}^m$. The aim is then to retrieve the signal $x$ from the observations $y$ \citep{CIBOOK1,CIBOOK2}. In the following subsections, the main advances in solving imaging inverse problems are reviewed 
and the relevant background on Bayesian multi-fidelity models is covered.

\subsection{Linear models and Hand-crafted Priors}

In many CI problems, the forward observation model can be approximately described as a linear operator $A \in \mathbb{R}^{m \times n}$ and some independent noise $\epsilon \in \mathbb{R}^m$ \citep{CIBOOK1,CS1}, such that the measurements $y$ are assumed to be generated as

\begin{equation}\label{linsys}
	y = Ax + \epsilon.
\end{equation}

The noise $\epsilon$ often follows simple statistics, such as Gaussian or Bernoulli distributions, depending on the instruments and imaging settings. This choice of forward model is computationally advantageous for retrieval algorithms, as it can be run efficiently through a simple linear projection, and is often a sufficiently good approximation to the ``true" observation process.
The difficulty in retrieving the observed signal $x$ from an observation $y$ in this context derives from the fact that in CI inverse problems the operator $A$ is often poorly conditioned and consequentially the resulting inverse problem is ill-posed. Put differently, the inverse $A^{-1}$ is not well-defined and small errors in $y$ result in large errors in the naive estimation $x \simeq A^{-1}y$. To overcome this issue, the classical approach is to formulate certain prior assumptions about the nature of the target $x$ that help in regularising its retrieval.

\subsubsection{Maximum a Posteriori Inference}

A widely adopted framework is that of maximum a posteriori (MAP) inference with analytically defined prior assumptions. The aim is to find a solution which satisfies the linear observations well, while imposing some properties which the target $x$ is expected to retain. Under Gaussian noise assumptions, the estimate of $x$ is recovered by solving a minimisation problem of the form 
\begin{equation}\label{linobj}
	\underset{x}{\arg\min} \quad \frac{1}{2}||Ax-y||^2 + \lambda h(x),
\end{equation}
where $|| \cdot ||$ indicates the Euclidean norm, $\lambda$ is a real positive parameter that controls the weight given to the regularisation and $h(x)$ is an analytically defined penalty function that enforces some desired property in $x$. In the case of images, it is common to assume that $x$ is sparse in some basis, such as frequency or wavelets, leading to $\ell_1$-norm penalty functions
\citep{CS1,CS2,SPARSEREC1}. For such choices of penalty, and other common ones, the objective function of equation \ref{linobj} is convex. This makes the optimisation problem solvable with a variety of efficient methods \citep{CS1,PRIMALDUAL} and provides theoretical guarantees on the recoverability of the solution \citep{CSO}. 

The aforementioned framework has been widely applied to perform CI. For instance, many image restoration tasks, such as de-blurring, up-sampling and in-painting have been formulated as ill-conditioned linear inverse problems and are solved as described above \citep{REGIM}. Various more complex sensing models can also be cast as linear operators, leading to the use of constrained optimisation in several CI systems that rely on ill-posed observations, such as sparse CT \citep{CT1}, single pixel photography \citep{SP2} and imaging of objects hidden from view \citep{CORNER}.

\subsubsection{Bayesian Inference}

MAP inference aims at recovering a single optimal solution to a given inverse problem. While such retrieval is arguably useful in many settings, it is not a complete description of the solution space. For a given ill-posed inverse problem there may be many solutions that satisfy similarly well the observed measurements and the prior assumptions. To capture the variability of such solution spaces, hence implicitly estimating reconstruction errors, the inverse problem can be cast as a Bayesian inference task; given an observation likelihood $p(y|x)$ and a signal prior $p(x)$, the aim is to retrieve the posterior PDF of solutions 
$p(x|y) = p(y|x)p(x)/p(y)$.
Estimating the full distribution of solutions $p(x|y)$ is generally much harder than simply finding its maximum through MAP inference. 

Approximate inference for the aforementioned problem has been approached in different ways. A popular class of methods in settings of limited dimensionality is that of inference through Markov chain Monte Carlo (MCMC) processes, with different choices of conditional sampling having been proposed \citep{MCMC_REVIEW,BCS,MCMC1,MCMC2,MCMC3}. 
Despite their guarantees of convergence to accurate approximations, MCMC methods are often prohibitively expensive for CI problems, as images are rather high dimensional. A second class of approaches is that of variational inference. These methods aim to use a tractable parametric PDF to approximate the true posterior $p(x|y)$ \citep{BINV, BINV2}. 
Though they do not provide the same guarantees as MCMC methods, these approaches are typically more efficient and have been explored with different PDFs and optimisation techniques in the context of CI \citep{BDECONV,BSR,BDECONV2}.

\subsection{Machine Learning for Computational Imaging}

The increasing availability of data sets and continuous advancements in learning inference models enabled new possibilities for CI and inverse problems in general. Learning from real examples allows to derive retrieval models directly from empirical experience, instead of relying on analytically defined priors and observation processes. 
The main classes of machine learning methods for CI are reviewed below.

\subsubsection{Learning Inverse Mappings}

Most learning approaches can arguably be described as inverse mapping models; with enough example pairs available, a neural network can be trained to directly recover a signal $x$ from observations $y$ \citep{NNIP}. Many neural architectures of this type have been developed to perform different image reconstruction tasks. In particular, convolutional neural networks are popular choices due to their ability to capture local pixel dependencies in images \citep{CNNIR2,CNNCIR}. These models are trained solely with paired examples of observed targets $X^*$ as outputs and corresponding observations $Y^*$ as inputs. These target-observation pairs are collected from either experimental acquisitions or numerical simulations. Once the model is trained, a new empirical observation $y_j$ can be mapped to the corresponding target reconstruction estimate $x_j$ \citep{NNIP}. 


Directly learning inverse mappings retains a number of advantages compared to analytical methods. First, the model is trained with a set of images the target solution is assumed to belong to, implicitly making the signal assumptions more specific than, for example, sparsity in some basis. Second, the observation model $f(x)$ is not constrained to be linear, or even differentiable; so long as a large number of signal-observations pairs is available the network can be trained to perform the inversion. Third, once the model is trained, inference is non-iterative and thus typically much faster, allowing elaborate imaging systems to retrieve reconstructions in real time and even at video rate. 

State of the art performance has been demonstrated with specifically designed neural networks models in many common image processing tasks, such as deconvolution and super-resolution \citep{NNDECONV,NNSUPER}, as well as signal recovery from under determined projective measurements \citep{NNCS1}. 
However, learned inverse mappings for CI retain two main problems. The first is that their accuracy of inference is entirely dependent on the available training targets and observations, leading to the need of carrying out lengthy data collections or numerical simulations to ensure robustness. The second is that it is difficult to asses the reliability of a given reconstruction; the trained neural network returns a deterministic estimate of the target that is usually in the range of the training examples, making it difficult to recognise unsuccessful recovery. 

\subsubsection{Iterative Inference with Learned Priors}

A second class of learning methods that is conceptually closer to analytical techniques is that of MAP inference with learned prior knowledge. The general idea is to exploit a differentiable analytic observation model to maximise the agreement with recorded observations, as in traditional MAP inference, but build the regularising prior empirically, learning from examples of expected signals \citep{CSGEN1}. The prior assumptions can be captured and induced in different ways. One option is to train a function $H(x)$ to quantify how much a target $x$ is expected to belong to a given set of training examples. The solution is then found by iteratively solving the minimisation problem 

\begin{equation}\label{mapml}
	\underset{x}{\arg\min} \quad \frac{1}{2}||Ax-y||^2 + \lambda H(x),
\end{equation}
where $A$ is a linear operator describing the observation process. Different choices of function $H(x)$ have been explored in recent works. One such choice is to train a discriminator $D(x)$ to recognise targets which belong to the training class and then setting $H(x) = s(\log(D(x)))$, where $s(\cdot)$ is a Sigmoid function \citep{ONE}. One other popular choice is to train a de-noising function $N(x)$ on the set of expected targets and then use the distance between a target and its de-noised equivalent $H(x) = ||x-N(x)||$ \citep{MAP_PLUG1,MAP_PLUG3,MAP_PLUG2}. Machine learning has also been implemented to train optimisation methods to solve the minimisation of equation \ref{mapml}. In fact, the iterative update of the solution $x$ through the optimisation procedure in these settings is often interpreted as a recurrent neural network \citep{CS_RNN1,CS_RNN2}. In such a way, the iterative inference precision is empirically adjusted to the specific inversion task, hence gaining in efficiency and accuracy \citep{CS_RNN3}.

A second framework to infer learned properties in iterative MAP inference is that of constrained minimisation with generative models. In these techniques, a generative model, such as a generative adversarial network (GAN) or a variational auto-encoder (VAE), is trained with a data set of expected targets, resulting in a generator $G(z)$ that can synthesise artificial examples $x$ in the range of interest from low-dimensional latent noise variables $z$. The solution target is then assumed to be synthesised by such generator, resulting in the following minimisation

\begin{equation}\label{mapgen}
	\underset{z}{\arg\min} \quad \frac{1}{2}||A \cdot G(z)-y||^2.
\end{equation}

In such a way, the solution is constrained to be within the domain of the generative model, as the recovered $x$ is by definition generated from $z$, while at the same time agreement to the measurements is induced by minimising the distance to the observations $y$. Iterative inference with generative models has been demonstrated for linear observation processes and phase-less linear observation processes \citep{CSGEN1, CSGEN3,CSGEN2}.

MAP inference with learned prior methods do eliminate the problem of data collection, as training is performed using solely examples of targets, while the nature of observations is incorporated through an analytically defined model \citep{ONE}. However, compared to learning inverse mappings, it comes with significant drawbacks. Firstly, the target-observations relationship is described entirely by an analytical model, sacrificing the desirable ability of machine learning to generalise mappings from empirical evidence. Secondly, these methods infer a solution to an inverse problem iteratively, excluding real time reconstruction applications. Furthermore, like learned inverse mappings, the solutions returned are deterministic, hence making it difficult to assess the reliability of a reconstruction. 

\subsubsection{Conditional Generative Models}

A promising direction to overcome the reliability problem is that of conditional generative models; instead of learning a deterministic mapping from observations to single reconstructions, a generative model is trained to generate different targets conditioned on given observations. The generation of multiple solutions from the same measurements can be probabilistically interpreted as sampling from the recovered posterior distribution. From these samples, uncertainty metrics, such as mean and standard deviation, can be inferred.
The recovered uncertainty can then be used to asses the reliability of a particular reconstruction or be propagated to automated decisions.
Recent advances in variational methods and adversarial models allow to train efficiently approximate inference through generative models that scale to the dimensionalities and numbers of examples typically needed for imaging tasks \citep{VAE,PIXVAE,CGAN}. Building upon these advancements, different conditional generative models have been developed in recent years, with the most commonly adopted being conditional Generative Adversarial Networks (CGANs) and conditional variational auto-encoders (CVAEs) \citep{CGAN2,CVAE,PLUG}.

Conditional generative models have been applied to perform a range of inference tasks, such as classification \citep{CVAE}, generation conditioned on classes \citep{PLUG,GAN_CC}, image-from-text inference \citep{GAN_TI,CVAET} and missing value imputation \citep{GAN_MVI,MVI}. However, within CI, they have been largely restricted to inference from simple deterministic observation, such as missing pixels or down-sampling \citep{PLUG,ITRA}, with the exception of recent work by \cite{AH}, in which a specifically designed GAN model is used to retrieve medical images from CT scans. In fact, the direct application of conditional generative models in CI is challenging because of the large data volumes requirements. Conditional generative models, in their common form, need a large number of object-condition pairs to train upon. In CI settings, this translates to the need of obtaining a large number of sufficiently accurate target-observation examples, which are unique to imaging instruments and hence expensive to collect.

\subsubsection{Semi-Supervised Conditional Generative Models}

Another closely related extension of generative models is that of semi-supervised learning with generative models. Similarly to conditional generative models, these methods introduce conditions on their generations, but are able to train with data sets where conditions are only available for a portion of the examples \citep{SEM_GEN_1,SEM_GEN_2,SEM_GEN_3}. They achieve this by introducing auxiliary models that map inputs to conditions and are trained jointly with the generator.

In some sense, the presented framework belongs to this class of methods, as the forward model component plays an analogous role to the auxiliary model in these systems, but retains two main distinctions:\vspace{-5pt}
\begin{enumerate}[i]
  \setlength{\itemsep}{1pt}
  \setlength{\parskip}{0pt}
  \setlength{\parsep}{0pt}
\item  The forward model is built as a multi-fidelity process that incorporates physical observation models and therefore exploits domain expertise to infer conditions (measurements in our setting). 
\item  Instead of training the auxiliary model and the conditional generator jointly, the proposed method applies a two step procedure, as shown in figure \ref{fig:models_2}. 
\end{enumerate}\vspace{-5pt}

The reason for the latter difference is that one needs to ensure synthetic measurements do not contain more information about the targets than the real ones. In the imaging setting, training the forward and reconstruction models jointly encourages the former to ``hide" additional information about the targets in the generated observations that real ones do not possess. In common semi-supervised settings, this is generally not a problem; the conditions, or auxiliary information, tend to be very low dimensional, e.g. labels, and the arising information bottleneck naturally prevents this effect. This is not true in imaging settings, where recorded measurements easily have more dimensions than the targets.

\subsection{Multi-Fidelity Bayesian Models}

Multi-fidelity methods exploit both highly accurate but expensive data and less accurate but cheaper data to maximise the accuracy of model estimation while minimising computational cost \citep{MF}. In multi-fidelity Bayesian inference, the most accurate predictions, or high-fidelity outputs, are considered to be draws from the underlying true density of interest and the aim is to approximately recover such high-fidelity outputs from the corresponding inputs and low-fidelity outputs of some cheaper computation \citep{MFI}. Within Bayesian approaches to solve inverse problems, multi-fidelity models have been used to minimise the cost of estimating expensive forward processes, in particular with MCMC methods to efficiently estimate the likelihood at each sampling step \citep{MF_MCMC}.

In Bayesian optimisation settings, the difference between high and low fidelity predictions is commonly modeled with Gaussian processes, where approximate function evaluations are made cheap by computing low-fidelity estimates and subsequently mapping them to high-fidelity estimates with a Gaussian process \citep{MF_GP,MF_GP2}. In CI settings, Gaussian process multi-fidelity models are difficult to apply, as the available volume of data and the dimensionality of the targets and observations are potentially very large. Recent work by \cite{MF_VAE} proposes to model high-fidelity data with conditional deep generative models, which are instead capable of scaling to the volumes and dimensionalities needed in imaging applications. The multi-fidelity component of the framework presented here follows these ideas and exploits a deep CVAE to model high fidelity data when inferring an accurate forward observation process.
      
\section{Variational Framework for Imaging Inverse Problems} \label{sec:technical}

Differently from previous approaches, the proposed variational framework is built to learn from all the useful data and models typically available in CI problems. In the following subsections the problem of Bayesian learning in this context is defined and the components of the proposed variational learning method are derived and motivated.




\subsection{Problem Description}


\subsubsection{The Bayesian Inverse Problem} \label{problem_description}

The aim of computational imaging is to recover a hidden target $x_j \in \mathbb{R}^{N}$ from some associated observed measurements $y_{j} \in \mathbb{R}^{M}$. In the Bayesian formulation, the measurements $y_{j}$ are assumed to be drawn from an observation distribution $p(y|x_j)$ and the objective is to determine the posterior $p(x|y_j)$; the distribution of all possible reconstructions. Following Bayes' rule, the form of this posterior is 
\begin{equation}\label{bayes2}
	p(x|y_j) = \frac{p(y_j|x)p(x)}{p(y_j)}.
\end{equation}
The observation distribution $p(y|x)$, often referred to as the data likelihood, describes the observation process, mapping targets to measurements. Given any ground truth target $x_i$ the corresponding measurements that are physically recorded $y_i$ are draws from the data likelihood $y_i \sim p(y|x_i)$. The prior distribution $p(x)$ models the assumed knowledge about the targets of interest. This PDF is the distribution of possible targets prior to carrying out any measurement. Finally, the marginal likelihood $p(y) = \int p(x) p(y|x) dx$ is the distribution of all possible measurements $y$.
The goal of variational inference is to learn a non-iterative approximation to the true intractable posterior distribution of equation \ref{bayes2} for arbitrary new observations $y_j$. That is, learning a parametric distribution $r_{\theta}(x|y)$ which well approximates the true posterior $p(x|y)$ for any new observation $y_j \sim p(y)$ and from which one can non-iteratively draw possible reconstructions $x_{j,i} \sim r_{\theta}(x|y_j)$.

\subsubsection{Information Available}

For a given imaging inverse problem of the type described above, there are generally three main sources of information that can be exploited to obtain the best estimate of the target posterior. The first is empirical observations. Physical experiments to collect sets of ground-truth targets $X^* \in \mathbb{R}^{N \times K}$ and associated observations $Y^* \in \mathbb{R}^{M \times K}$ can be recorded with the imaging apparatus of interest. The number of these acquisitions $K$ is normally limited by the time and effort necessary for experimental preparation and collection, or alternatively by computational cost, if these are obtained through numerical simulations. However, empirical target-observation pairs are the most accurate evaluation of the true observation process and therefore can be very informative. A recorded observation $y_k \in Y^*$ obtained when imaging a target $x_k \in X^*$ can be interpreted as a sample from the true data likelihood $y_k \sim p(y|x_k)$.

The second source of information is domain expertise. The measurement process, mapping targets to observations, is described by a physical phenomenon. With knowledge of such phenomenon, one can construct a functional mapping, normally referred to as forward model, which computes observations' estimates $\widetilde{y}$ from targets $x$. For instance, many observation processes in CI settings can be approximately modelled by a linear transformation and independent Gaussian or Poisson noise \citep{CS1}. It is clearly infeasible to obtain analytical models that perfectly match reality. However, an analytical forward model can provide inexpensive approximations $\widetilde{y}$ to the true observations $y$ that can be computed for any target $x$. In the Bayesian formulation, a forward model can be interpreted as a closed form approximation $p(\widetilde{y}|x)$ to the true data likelihood $p(y|x)$.

Lastly, many examples of the targets of interest $X \in \mathbb{R}^{N \times L}$ are often available in the form of unlabelled data sets. Because collection of this type of data is independent of the imaging apparatus, the number of available examples $L$ is expected to be much greater than the number of empirical acquisitions $K$. In fact, many large image data sets containing relevant targets for CI applications are readily available and easily accessible. These target examples $x_l \in X$ can be interpreted as draws from the prior distribution $x_l \sim p(x)$. In summary, the available sources of information are

\begin{itemize}
    \item Limited sets of ground-truth targets $X^* = \{x_{k=1:K}\}$ and associated observations $Y^* = \{y_{k=1:K}\}$, the elements of which are point samples of the true data likelihood $y_k \sim p(y|x_k)$.
    \item An analytical forward model providing a closed form approximation for the true data likelihood $p(\widetilde{y}|x) \approx p(y|x)$.
    \item A large set of target examples $X = \{x_{l=1:L} \}$ corresponding to prior samples $x_l \sim p(x)$, where $L \gg K$.
\end{itemize}

The scope of this paper is to design a framework for learning the best possible approximate distribution $r_{\theta}(x|y)$ by exploiting all the available sources of information described above.

\subsection{Multi-Fidelity Forward Modelling} \label{multi}

Before training the inversion, an approximate observation distribution $p_{\alpha}(y|x)$ is trained to fit the true data likelihood $p(y|x)$. Learning this observation distribution first allows effective incorporation of domain expertise, as in CI problems this is usually available in the form of an analytical forward model. Furthermore, training the observation model is expected to require far fewer training input-output pairs than training the corresponding inversion, as most forward models are well-posed, while the corresponding inverse problems are often ill-posed. A good approximation to the data likelihood $p_{\alpha}(y|x)$ can therefore be learned with a much lower number $K$ of experimental ground-truth targets $X^*$ and measurements $Y^*$ than would be required to train a good approximate posterior $r_{\theta}(x|y)$ directly.

In order to make use of the analytical approximation $p(\widetilde{y}|x)$, hence incorporating domain expertise in the training procedure, the approximate observation distribution is chosen as
\begin{equation}\label{fm1}
	p_{\alpha}(y|x) = \int p(\widetilde{y}|x)p_{\alpha}(y|x,\widetilde{y})d \widetilde{y}.
\end{equation}
In such a way, the inference of a measurement $y$, a high-fidelity prediction, from a target $x$ can exploit the output $\widetilde{y}$ of the analytical forward model $p(\widetilde{y}|x)$, which instead is considered a low-fidelity prediction. The parametric component to be trained is then the conditional $p_{\alpha}(y|x,\widetilde{y})$, which returns high-fidelity sample measurements $y$ from targets $x$ and low-fidelity predictions $\widetilde{y}$.

To provide flexible inference in the general case, the PDF $p_{\alpha}(y|x,\widetilde{y})$ is chosen to be a latent variable model of the form
\begin{equation}\label{fm2}
	p_{\alpha}(y|x,\widetilde{y}) = \int p_{\alpha_1}(w|x,\widetilde{y})p_{\alpha_2}(y|x,\widetilde{y},w)dw.
\end{equation}
The two parametric distributions $p_{\alpha_1}(w|x,\widetilde{y})$ and $p_{\alpha_2}(y|x,\widetilde{y},w)$ are chosen to be Gaussian distributions, the moments of which are outputs of neural networks with weights $\alpha_1$ and $\alpha_2$ respectively.\footnote{The distribution $p_{\alpha_2}(y|x,\widetilde{y},w)$ can alternatively be chosen to match some other noise model if the observation noise is known to be of a particular type, such as Poisson or Bernoulli.} 
The model of equation \ref{fm1} is then trained to fit the sets of experimental ground-truth targets and measurements $X^*$ and $Y^*$, as these are point samples of the true data likelihood of interest $y_k \sim p(y|x_k)$. The optimisation to be performed is the log likelihood maximisation
\begin{equation}\label{fm3}
	\underset{\alpha_1,\alpha_2}{\arg\max} \quad \log p_{\alpha}(Y^*|X^*) = \sum_{k=1}^K \log \int p(\widetilde{y}|x_k)\int p_{\alpha_1}(w|x_k,\widetilde{y})p_{\alpha_2}(y_k|x_k,\widetilde{y},w)dw d \widetilde{y}.
\end{equation}
Due to the integration over latent variables $w$, the maximisation of equation \ref{fm3} is intractable to directly perform stochastically. However, problems of this type can be approximately solved efficiently with a variational auto-encoding approach, in which a parametric recognition model is used as a sampling function  \citep{VAE, CVAE}. The VAE formulation for the multi-fidelity model is presented in detail in supplementary section \ref{VAE_for}. Through this approach, training of the parameters $\alpha = \{ \alpha_1, \alpha_2 \}$ and $\beta$ can be performed through the following stochastic optimisation:
\begin{eqnarray}\label{fm6}
\begin{split}
	\underset{\alpha_1,\alpha_2, \beta}{\arg\max} \quad \sum_{k=1}^K \sum_{v=1}^V \Bigg[ \sum_{s=1}^S \log p_{\alpha_2}(y_k|x_k,\widetilde{y}_v,w_s) &- D_{KL}(q_{\beta}(w|x_k,y_k,\widetilde{y}_v) || p_{\alpha_1}(w|x_k,\widetilde{y}_v)) \Bigg].
\end{split}
\end{eqnarray}
This bound is maximised stochastically by drawing samples from the approximate distribution $p(\widetilde{y}|x)$ and subsequently from a recognition model $q_{\beta}(w|x_k,y_k,\widetilde{y})$, which is chosen as an isotropic Gaussian distribution, the moments of which are outputs of a neural network taking as inputs targets $x$, high-fidelity measurements $y$ and low-fidelity measurements $\widetilde{y}$.

\begin{figure*}[t!]
  \centering
  \includegraphics[width=\linewidth]{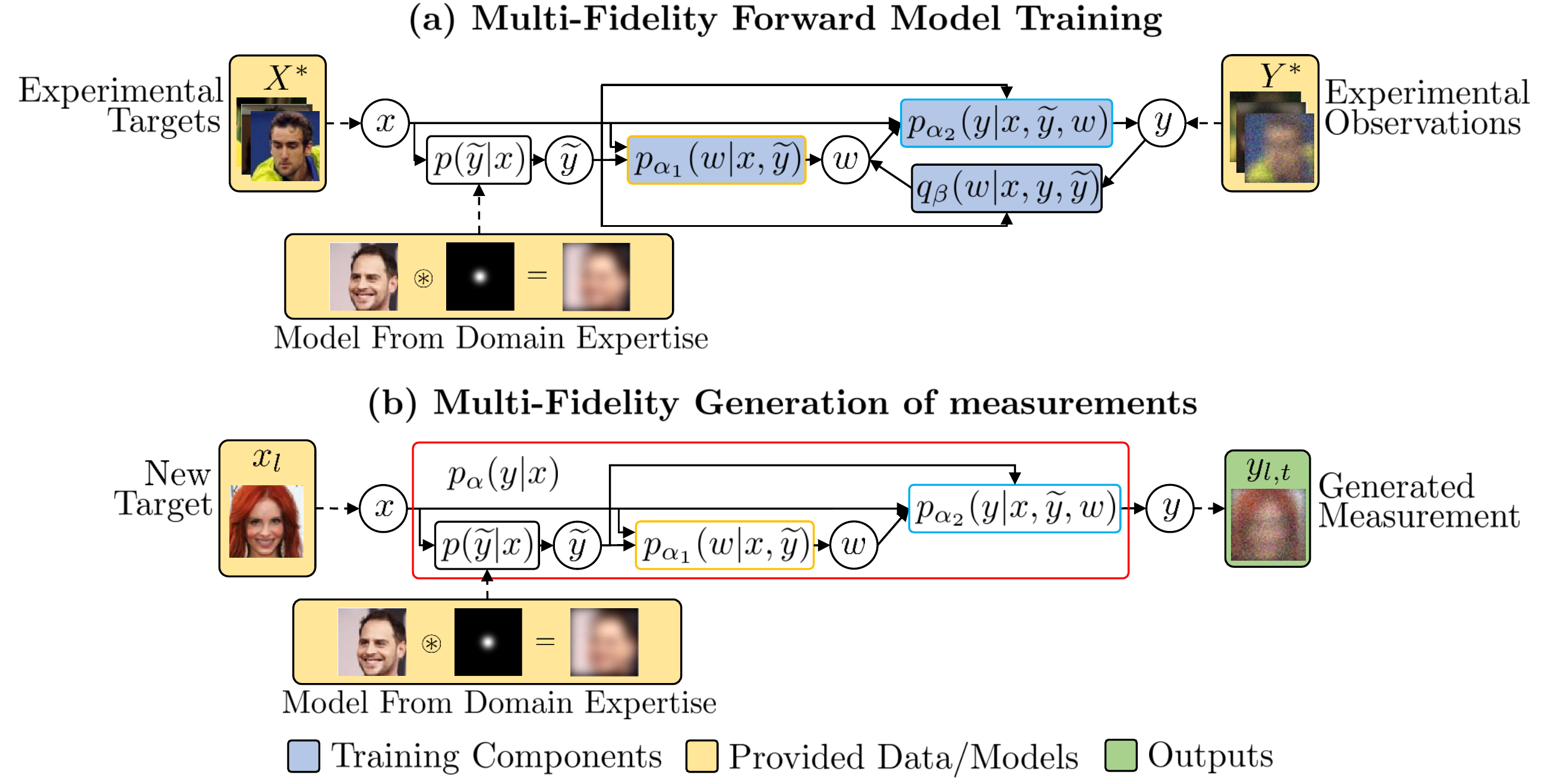}
\caption{Multi-fidelity forward modelling. (a) The two conditional distributions $p_{\alpha_1}(w|x,\widetilde{y})$ and $p_{\alpha_2}(y|x,\widetilde{y},w)$, parametric components of the multi-fidelity forward model $p_{\alpha}(y|x)$, are trained with an auto-encoding approach, making use of a recognition model $q_{\beta}(w|x,y,\widetilde{y})$. These distributions are trained with an analytical forward model defining $p(\widetilde{y}|x)$, experimental ground-truth targets $X^*$ and corresponding observations $Y^*$. (b) once the parameters $\alpha = \{\alpha_1, \alpha_2 \}$ have been trained, the learned distributions can be used to generate multi-fidelity estimates of observations $y_{l,t}$ from a new target $x_l$. First, a low fidelity estimate $\widetilde{y}_v$ is generated through the analytical observation model $p(\widetilde{y}|x_l)$. Second, this estimate and the corresponding target are used to draw a latent variable from $p_{\alpha_1}(w_s|x_l,\widetilde{y}_v)$. Third, the target $x_l$, low-fidelity estimate $\widetilde{y}_v$ and latent variable $w_s$ are used to generate a high-fidelity observation's estimate $y_{l,t}$ by sampling from $p_{\alpha_2}(y|x_l,\widetilde{y}_v,w_s)$. Performing these operations in sequence corresponds to running the multi-fidelity forward model $y_{l,t} \sim p_\alpha(y|x)$.}
\label{fig:multifidelity}
\end{figure*}

Sampling from the approximate likelihood $\widetilde{y}_v \sim p(\widetilde{y}|x_k)$ is equivalent to running the analytical forward observation model. For instance, in the case of a linear observation model, the samples $\widetilde{y}_v$ are computed as $\widetilde{y}_v = A x_k + \epsilon_v$, where $A$ is the linear mapping given by the model and $\epsilon_v$ is drawn from the noise process characteristic of the apparatus of interest. Pseudo-code for the multi-fidelity forward model training is in supplementary section \ref{alg}.

Once the weights $\alpha$ have been trained through the maximisation of equation \ref{fm6}, it is possible to inexpensively compute draws $y_{l,t}$ from the multi-fidelity data likelihood estimate $p_{\alpha}(y|x_l)$ given a new target $x_l$ as
\begin{equation}\label{fm7}
	y_{l,t} \sim p_{\alpha_2}(y|x_l,\widetilde{y}_v,w_s), \quad  \text{where} \quad  \widetilde{y}_v \sim  p(\widetilde{y}|x_l) \quad \text{and} \quad w_s \sim p_{\alpha_1}(w|x_l,\widetilde{y}_v).
\end{equation}
Computing a forward model estimate with the trained multi-fidelity likelihood consists of three consecutive computations. First, a low-fidelity estimate $\widetilde{y}_v$ is computed by running the analytical forward model. Second, a latent variable $w_s$ is drawn from the latent distribution $p_{\alpha_1}(w|x_l,\widetilde{y}_v)$. Lastly, the high-fidelity measurement estimate $y_{l,t}$ is drawn from the conditional $p_{\alpha_2}(y|x_l,\widetilde{y}_v,w_s)$. As all of these operations are computationally inexpensive, running the resulting multi-fidelity forward model is also inexpensive.

\subsection{Variational Inverse Model}


To learn an inversion model, the approximate posterior distribution $r_{\theta}(x|y)$ is trained to recover targets from observations, exploiting the learned PDF $p_{\alpha}(y|x)$ to generate measurements from the large data set of target examples $X$. In such a way, training of the approximate posterior $r_{\theta}(x|y)$ can exploit the large number $L \gg K$ of target examples $X$, even though no corresponding measurements are available, as estimates of these are generated implicitly during training through the learned forward model $p_{\alpha}(y|x)$. sampling synthetic measurements from $p_{\alpha}(y|x)$ also introduces variation in the training inputs to $r_{\theta}(x|y)$, improving generalisation in a similar way to noise injection strategies \citep{NI}.

The aim of this training stage is to train a parametric distribution $r_{\theta}(x|y)$ to match the true posterior $p(x|y)$. To this end, the expectation of the cross entropy $H[p(x|y),r_{\theta}(x|y)]$ under the measurements' distribution $p(y)$ is minimised with respect to the model's variational parameters $\theta$,
\begin{eqnarray}\label{varapp}
\begin{split}
	\underset{\theta}{\arg\min} \enskip  \mathbb{E}_{p(y)} H[p(x|y),r_{\theta}(x|y)] =  
\underset{\theta}{\arg\max} \enskip  \mathbb{E}_{p(y)} \! \int p(x|y) \log r_{\theta}(x|y) dx.
\end{split}
\end{eqnarray}
The optimisation of Equation \ref{varapp} is equivalent to fitting $r_{\theta}(x|y)$ to the true posterior $p(x|y)$ over the distribution of measurements that are expected to be observed $p(y)$. This objective function can be simplified to give
\begin{eqnarray}\label{distCVAE}
\begin{split}
	 \mathbb{E}_{p(y)} \int p(x|y) \log r_{\theta}(x|y) dx &= 
	 \int \!\!\! \int p(y)\frac{p(y|x)p(x)}{p(y)} \log r_{\theta}(x|y) dx dy\\ &= \int p(x) \int p(y|x) \log r_{\theta}(x|y) dy dx. 
\end{split}
\end{eqnarray}
In order to stochastically estimate and maximise the expression of equation \ref{distCVAE}, drawing samples from the prior $x_l \sim p(x)$ and from the likelihood $y_{l,t} \sim p(y|x_l)$ needs to be realizable and inexpensive. In the case of the former, a large ensemble of samples is readily available from the data set of target examples $X$. Therefore, to approximately sample from the prior one only needs to sample from this data set. On the other hand, sampling from the likelihood $p(y|x_l)$ is not possible, as the form of the true forward observation model is not accessible. However, the previously learned multi-fidelity forward model $p_{\alpha}(y|x)$, described in subsection \ref{multi}, offers a learned approximation to the data likelihood from which it is inexpensive to draw realisations. The objective of equation \ref{distCVAE} to be maximised can then be approximated as
\begin{eqnarray}\label{distCVAEapp}
\begin{split}
	 \mathbb{E}_{p(y)} \int p(x|y) \log r_{\theta}(x|y) dx \simeq \int p(x) \int p_{\alpha}(y|x) \log r_{\theta}(x|y) dy dx. 
\end{split}
\end{eqnarray}
In this form, stochastic estimation is inexpensive, as prior samples $x_l \sim p(x)$ can be drawn from the data set $X$ and draws from the approximate likelihood $y_{l,t} \sim p_{\alpha}(y|x_l)$ can be computed by running the multi-fidelity forward model as described in subsection \ref{multi}.

\subsubsection{CVAE as Approximate Posterior}

The approximate distribution $r_{\theta}(x|y)$ needs to be of considerable capacity in order to accurately capture the variability of solution spaces in imaging inverse problems. To this end, the approximating distribution $r_{\theta}(x|y)$ is chosen as a conditional latent variable model
\begin{equation}\label{cgen}
\begin{split}
r_{\theta}(x|y) = \int r_{\theta_1}(z|y) r_{\theta_2}(x|z,y) dz.
\end{split}
\end{equation}
The latent distribution $r_{\theta_1}(z|y)$ is an isotropic Gaussian distribution $\mathcal{N}(z;\mu_z,\sigma^2_z)$, where its moments $\mu_z$ and $\sigma^2_z$ are inferred from a measurement $y$ by a neural network. The neural networks may be convolutional or fully connected, depending on the nature of the observed signal from which images need to be reconstructed. The likelihood distribution $r_{\theta_2}(x|z,y)$ can take different forms, depending on the nature of the images to be recovered and requirements on the efficiency of training and reconstruction. In the experiments presented here, the distribution $r_{\theta_2}(x|z,y)$ was set to either an isotropic Gaussian with moments determined by a fully connected neural network, taking concatenated $z$ and $y$ as input, or a convolutional pixel conditional model analogous to that of a pixelVAE \citep{PIXVAE}.

Latent variable models of this type have been proven to be powerful conditional image generators \citep{CVAE,PLUG} and therefore are expected to be suitable variational approximators for posteriors in imaging problems. With this choice of approximate posterior $r_{\theta}(x|y)$, the objective function for model training is
\begin{eqnarray}\label{max_lik}
\begin{split}
\underset{\theta_1, \theta_2}{\arg\max} \enskip  \int p(x) \int p_{\alpha}(y|x) \log \int r_{\theta_1}(z|y) r_{\theta_2}(x|z,y) dz dy dx.
\end{split}
\end{eqnarray}
As for the likelihood structure in the multi-fidelity forward modelling, directly performing the maximisation of equation \ref{max_lik} is intractable due to the integral over the latent space variables $z$. However, using Jensen's inequality, a tractable lower bound for this expression can be derived with the aid of a parametric recognition model $q_{\phi}(z|x,y)$.

\begin{figure*}[t!]
  \centering
  \includegraphics[width=\linewidth]{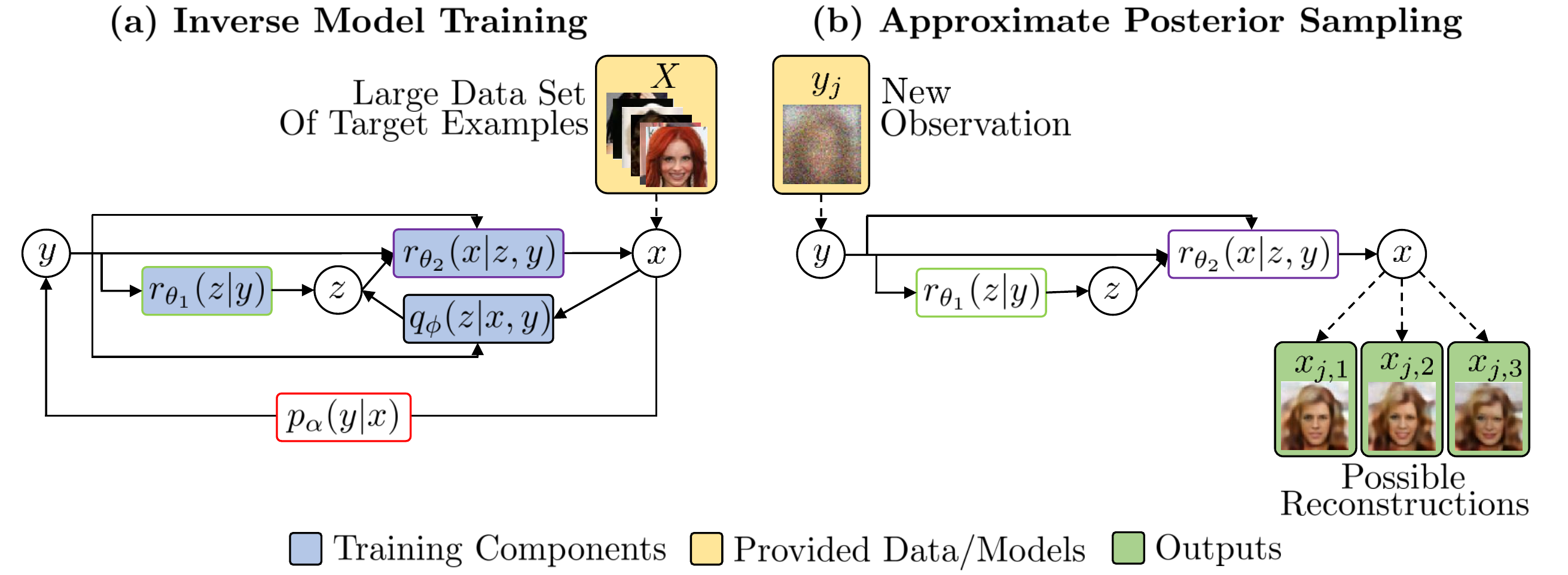}
\caption{Variational inverse model. (a) The model is trained to maximise the evidence lower bound on the likelihood of targets $x$ conditioned on observations $y$. The posterior components $r_{\theta_1}(z|y)$ and $r_{\theta_2}(x|z,y)$ are trained along with the auxiliary recognition model $q_{\phi}(z|x,y)$. Instead of training on paired targets and conditions, as for standard CVAEs, the model is given target examples $X$ alone and generates training conditions $y$ stochastically through the previously learned multi-fidelity forward model $p_{\alpha}(y|x)$. (b) Given new observations $y_j$, samples from the approximate posterior $r_{\theta}(x|y_j)$ can be non-iteratively generated with the trained model by first drawing a latent variable $z_{j,i} \sim r_{\theta_1}(z|y_j)$ and subsequently generating a target $x_{j,i} \sim r_{\theta_2}(x|z_{j,i},y_j)$.}
\label{fig:inverse}
\end{figure*}

\begin{figure*}[t!]
  \centering
  \includegraphics[width=\linewidth]{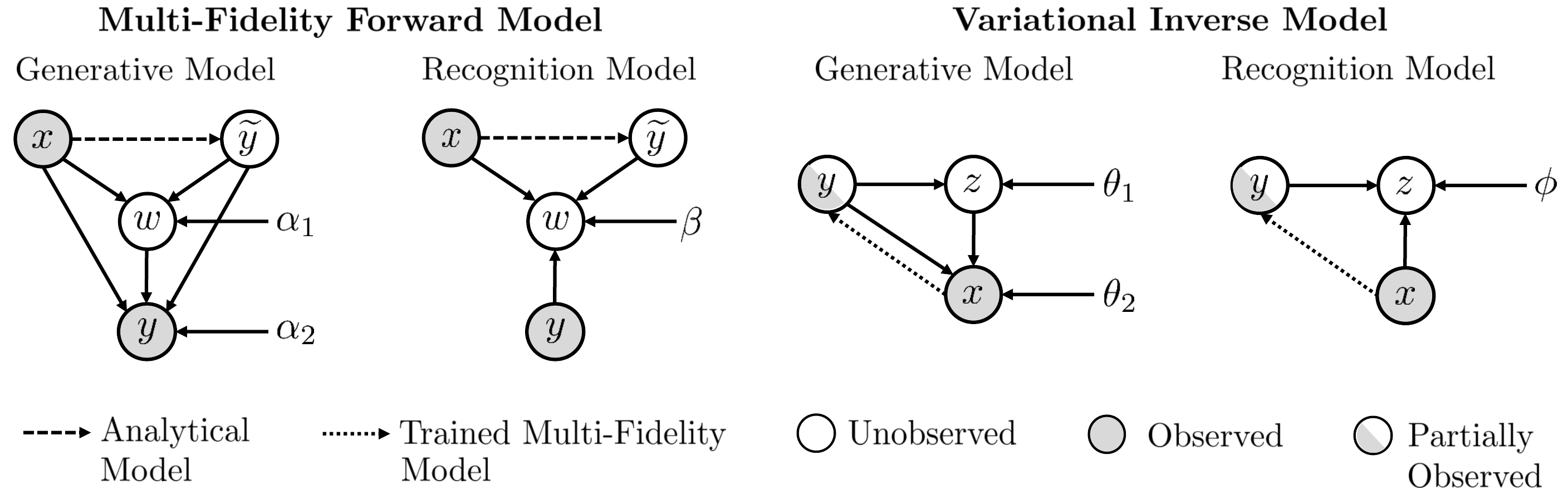}
\vspace{-0.7cm}
\caption{Graphical models for training of the multi-fidelity forward model and the variational inverse model.}
\label{fig:graphical_models}
\end{figure*}

As for the forward multi-fidelity model, the recognition model $q_{\phi}(z|x,y)$ is an isotropic Gaussian distribution in the latent space, with moments inferred by a neural network, taking as input both example targets $x$ and corresponding observations $y$. This neural network may be fully connected, partly convolutional or completely convolutional, depending on the nature of the targets $x$ and observations $y$. The VAE formulation for the Variational inverse problem is presented in detail in supplementary section \ref{VAE_inv}. Making use of this lower bound, we can define the objective function for the inverse model as
\begin{eqnarray}\label{VICI_obj}
\begin{split}
\underset{\theta_1, \theta_2, \phi}{\arg\max} \enskip  \sum_{l=1}^L \sum_{t=1}^T \left[\sum_{s=1}^S \log r_{\theta_2}(x_l|z_s,y_{l,t}) - D_{KL}(q_{\phi}(z|x_l,y_{l,t}) || r_{\theta_1}(z|y_{l,t})) \right],
\end{split}
\end{eqnarray}
where target examples are drawn from the large data set as $x_l \sim X$, measurements are generated with the multi-fidelity model as $y_{l,t} \sim p_{\alpha}(y|x_l)$ and latent variables are drawn from the recognition model as $z_s \sim q_{\phi}(z|x_l,y_{l,t})$, using the reparametrisation trick presented in \cite{VAE}. The variational approximate posterior $r_{\theta}(x|y)$ is trained by performing the maximisation of equation \ref{VICI_obj} through steepest ascent. The training procedure is schematically shown in Figure \ref{fig:inverse}(a) and detailed as a pseudo-code in supplementary \ref{alg}. The models employed during training of the multi-fidelity forward model and the variational inverse model are both summarised in the graphical models of figure \ref{fig:graphical_models}.

\subsubsection{Inference}

Once the variational parameters $\theta = \{\theta_1, \theta_2\}$ have been trained, the learned approximate posterior can be used to generate draws $x_{j,i} \sim r_{\theta}(x|y_j)$ conditioned on new measurements $y_j$. Draws from the posterior are obtained by first drawing a latent variable $z_{j,i} \sim r_{\theta_1}(z|y_j)$ and subsequently generating a target $x_{j,i} \sim r_{\theta_2}(x|z_{j,i},y_j)$. Such generated samples can be interpreted as different possible solutions to the inverse problem and can be used in different ways to extract information of interest. For instance, one can compute per-pixel marginal means and standard deviations, in order to visualise the expected mean values and marginal uncertainty on the retrieved images. Figure \ref{fig:inverse}(b) schematically illustrates the approximate posterior sampling procedure.

It may be of interest to also estimate a single best retrieval $x_{j}^*$ given the observed measurements $y_j$, which would be the image yielding the highest likelihood $r_{\theta}(x_{j}^*|y_j)$. This retrieval can be performed iteratively, by maximising $r_{\theta}(x|y_j)$ with respect to $x$, as proposed by \cite{CVAE}. As the focus of this work is non-iterative inference, a pseudo-maximum non-iterative retrieval is instead used. Such retrieval is performed by considering the point of maximum likelihood of the conditional Gaussian distribution in the latent space $r_{\theta_1}(z|y_j)$, which is by definition its mean $\mu_{z,j}$. The pseudo-maximum reconstruction $x_j^*$ is then the point of maximum likelihood of $r_{\theta_2}(x|\mu_{z,j},y_j)$, which is also its mean $\mu_{x,j}$. This pseudo-maximum estimate adds the ability to retrieve an inexpensive near-optimal reconstruction, analogous to that recovered by deterministic mappings.

\section{Experiments}

The proposed framework is tested both in simulation and with real imaging systems. Firstly, quantitative simulated experiments are performed to compare the proposed training framework to other strategies using the same Bayesian networks. Secondly, the new approach is applied to phase-less holographic image reconstruction and imaging through highly scattering media, comparing reconstructions to the most recent state of the art in the respective fields.

\subsection{Simulated Experiments}

\begin{figure*}[t]
  \centering
  \includegraphics[width=\linewidth]{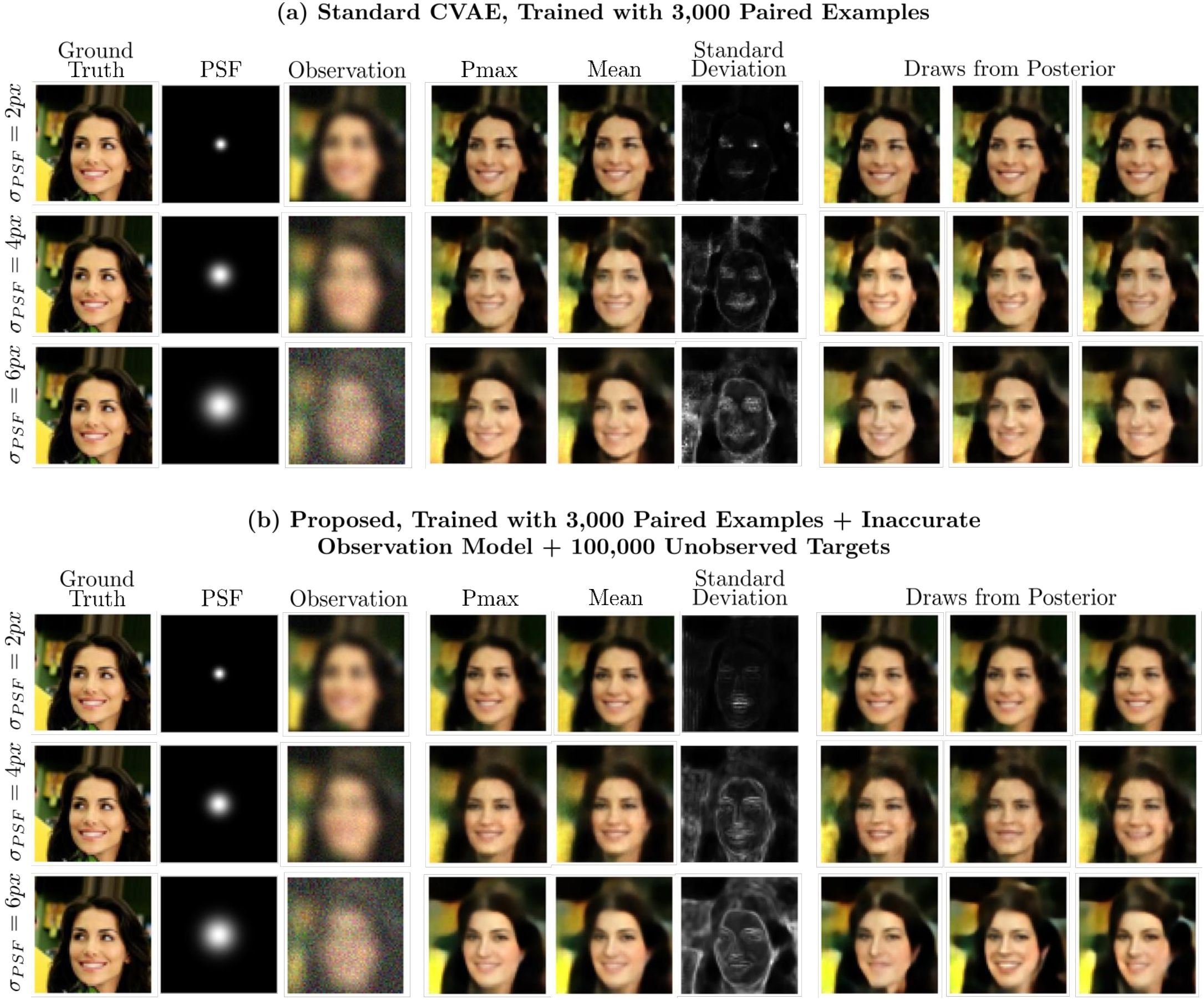}
\vspace{-0.7cm}
\caption{Comparison between standard CVAE trained with paired examples and proposed framework. (a) Posterior recovery obtained with a CVAE trained on $3,000$ available image-observation pairs. The number of available paired examples is not sufficient to train a CVAE capable of capturing the variability of the solution space and the model over-fits; draws from the posterior are all very similar independently of how ill-posed the de-convolution inverse problem is. (b) Posterior recovery obtained by training the CVAE with the proposed framework, exploiting all sources of information available. This model adequately captures the variability of the different solution spaces; as blurring and noise become more severe and the corresponding inverse problem more ill-posed, the draws from the posterior increasingly diversify, exploring the possible faces that could lie behind the associated observation.}
\label{fig:inc_fig}
\end{figure*}

Simple image restoration tasks are performed in simulation. Experiments include Gaussian blurring, down-sampling and partial occlusion with both CelebA and CIFAR examples \citep{celebA,CIFAR}. The test images are corrupted with the given transformation and additive Gaussian noise. Variational models are then used to perform reconstructions, with the aim of capturing the posterior of solutions to the resulting inverse problem.
To simulate typical CI conditions, only a small subset of images degraded with the true transformation is made accessible. However, the whole training set of ground truth images remains available, as this does not rely on the particular imaging instrument and can be sourced independently. In addition, an inaccurate degradation function is provided to simulate domain expertise. Inaccuracies compared to the true transformation are simulated with errors on the transformation's parameters. 

\subsubsection{Comparison with Standard training of CVAEs}

 As a first example, three different levels of Gaussian blurring and additive noise degradation conditions are considered with $64 \times 64$ images. The models are given $K = 3,000$ paired examples generated with the true transformation to train upon. The inaccurate observation model exploited by the proposed framework under-estimates the point spread function (PSF) width and noise standard deviation by $25\%$ compared to the true transformation. Reconstruction examples are shown in Figure \ref{fig:inc_fig}. More experimental details are given in appendix \ref{exp_1}.

\begin{figure*}[t]
  \centering
  \includegraphics[width=\linewidth]{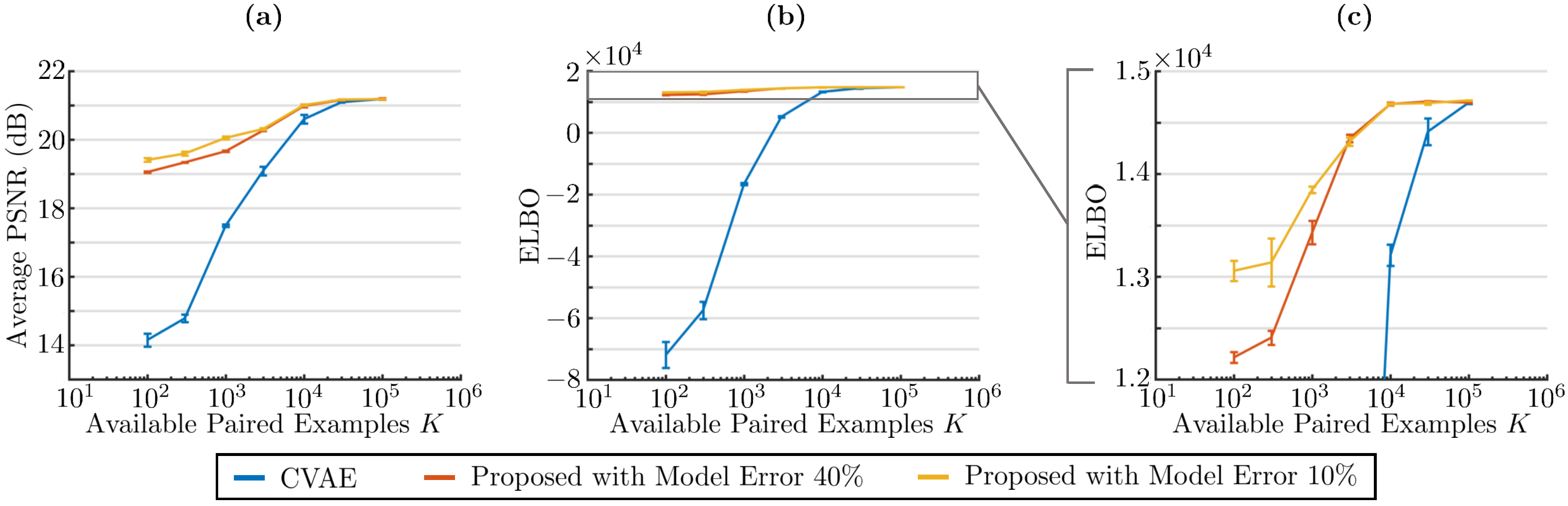}
\vspace{-0.7cm}
\caption{Posterior reconstruction from blurred CelebA images at varying number $K$ of paired training examples. (a) Average PSNR between reconstructed pseudo-maximum and ground truth images. (b-c) ELBO assigned to the test set by the trained models.
A standard CVAE requires a large number $K$ of paired images and observations to obtain accurate mean reconstructions (high PSNR values), and tens of thousands of examples before yielding distributions that approximately match the true posterior (high ELBO values). For many imaging tasks these would be prohibitively expensive to collect. By incorporating additional cheap sources of information, the proposed framework achieves high performance with far fewer paired examples, minimising data collection costs. Furthermore, training with a more accurate observation model yields superior performance when fewer paired examples are available, proving the ability of the framework to benefit from improved domain expertise.
}
\label{fig:Kvar}
\end{figure*}




With only $3,000$ image-observation pairs to train upon, the CVAE is not able to properly capture the variability of the solution space. The model returns compelling mean and pseudo-max reconstructions, but fails to explore the variation of possible solutions; different draws from the recovered posterior remain very similar to each other and do not properly represent the range of faces that could generate the observed blurred image. Contrarily, by including the additional data and model with the proposed framework, the CVAE is adequately trained. Draws from the posterior increasingly diversify as blurring and noise intensify, reflecting the increasing variance of the solution space. 

\begin{figure*}[t]
  \centering
  \includegraphics[width=\linewidth]{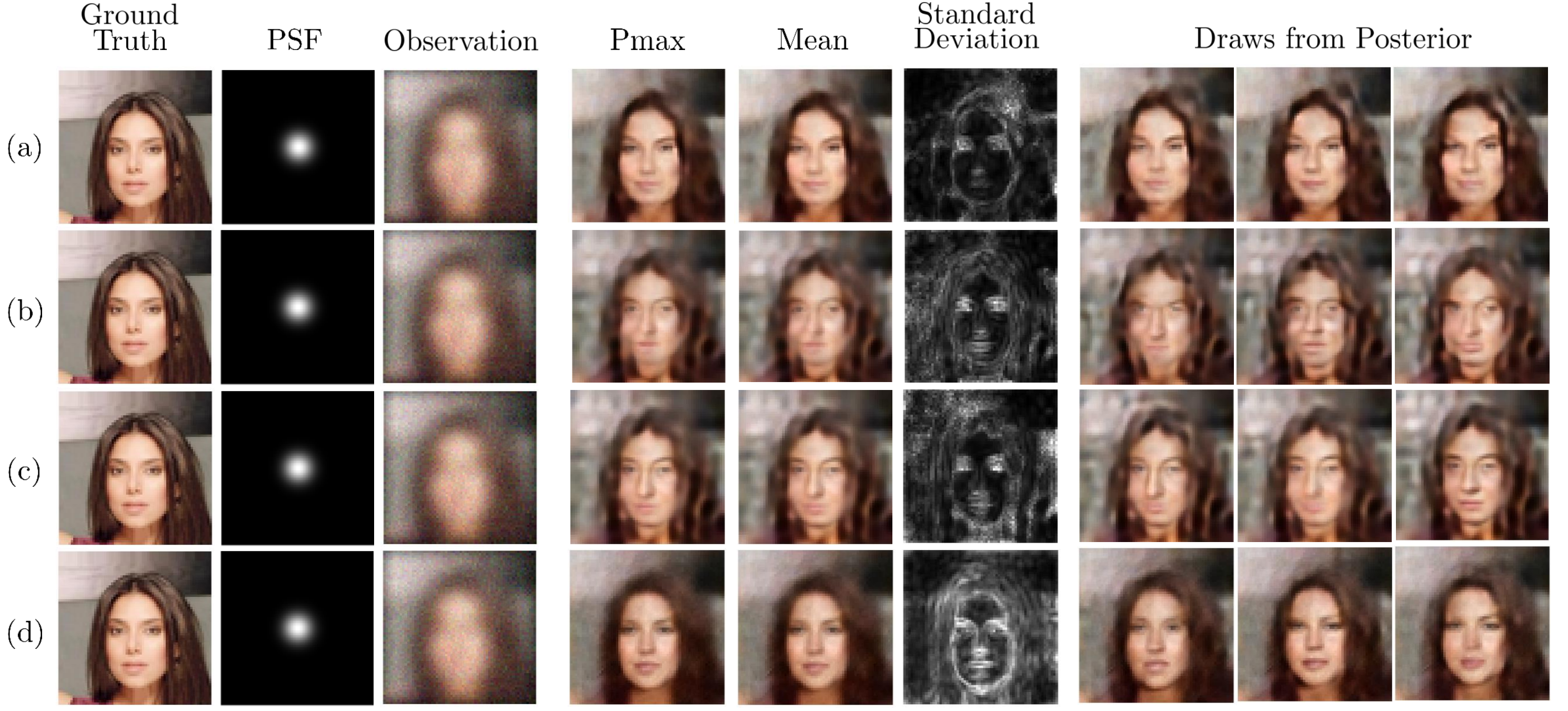}
\caption{Posterior recovery from blurred. (a) CVAE trained with $K = 3,000$ paired examples alone. With too few examples to train with, the model over-fits and draws from the posterior are all very similar. (b) CVAE trained with $L = 100,000$ target examples and corresponding simulated observations from the inaccurate observation model. Because the observation model does not match the true one encountered upon testing, reconstructions display noticeable artefacts. (c) CVAE trained with $K = 3,000$ paired examples in combination with $L = 100,000$ target examples and corresponding inaccurately simulated observations. The presence of real measurements in the training set improves reconstruction marginally, but artefacts are still largely present. (d) CVAE trained with proposed variational framework. The sources of information are exploited in a principled way, resulting in accurate posterior recovery; different draws explore various plausible reconstructions.
}
\label{fig:diff}
\end{figure*}

To test the proposed framework in different conditions, multiple experiments analogous to those illustrated in Figure \ref{fig:inc_fig}, with different relative model errors, are performed varying the number $K$ of available image-observation pairs. 
More experimental details are given in appendix \ref{exp_1}. As shown in Figure \ref{fig:Kvar},
a standard CVAE yields very low peak signal to noise ratio (PSNR) if the number $K$ of available paired training data is below a few thousands, indicating poor mean performance. The behaviour of the ELBO is even more dramatic, essentially suggesting complete inability to capture the posterior of solutions with less than a few tens of thousands paired examples. In many imaging settings, collecting such a high number of image-observation pairs would be extremely expensive. Instead, by incorporating additional cheap sources of information, the proposed framework displays appreciable PSNR and ELBO, even with very scarce paired image-observation examples. 
Furthermore, the use of a more accurate observation model was found to sensibly improve reconstructions at low numbers $K$ of available paired examples, to then converge towards similar performance as this was increased. Such results indicate that the proposed framework  is able to make better use of empirical data and domain expertise than the naive baselines;
the accuracy of the analytical observation model affects the recovery when availability of empirical evidence is low, but is progressively less influential as more data becomes available.

\begin{figure*}[t]
  \centering
  \includegraphics[width=\linewidth]{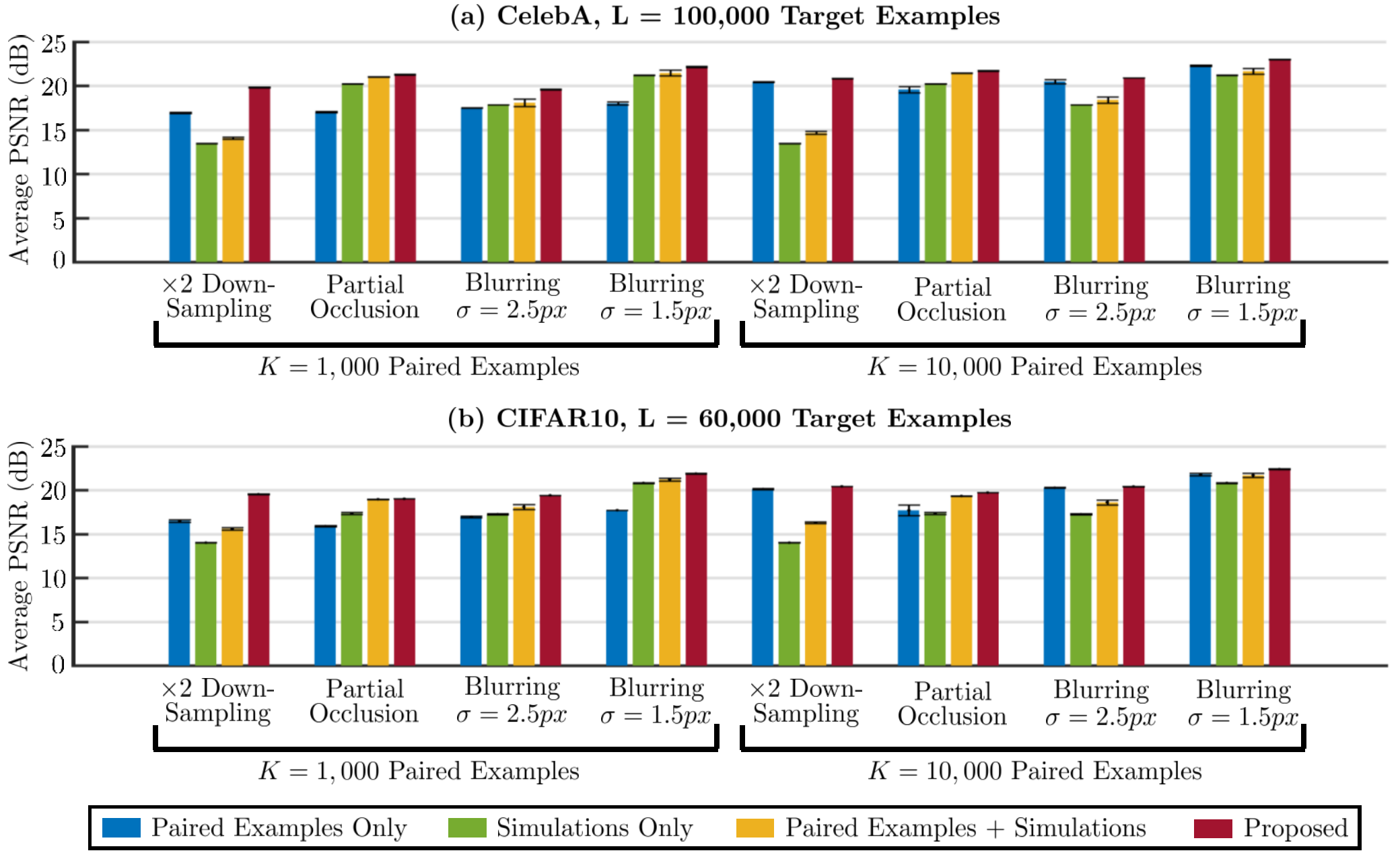}
\caption{average test PSNR between ground truth images and reconstructed pseudo-maxima for (a) CelebA and (b) CIFAR10 images. The proposed framework consistently outperforms other training methods that exploit part of or all of the same sources of information. 
}
\label{fig:psnr}
\end{figure*}

\subsubsection{Comparison with alternative Training Methods}

Given the small number $K$ of paired training data, a large number $L$ of target examples and an inaccurate observation model, one can conceive different naive ways to train a conditional generative model for inversion:\vspace{-\parskip}
\begin{enumerate}[i]
  \setlength{\itemsep}{1pt}
  \setlength{\parskip}{0pt}
  \setlength{\parsep}{0pt}
    \item Standard conditional training; discard the availability of target examples and domain expertise and train solely on $K$ empirical target-observation pairs.
    \item Use of domain expertise only; simulate a large number $L$ of measurements from all available targets through the analytical model and use these as pairs to train the model.
    \item Combining the previous two approaches; the $K$ targets for which empirical measurements are available are paired with them, while the $L$ unobserved targets are paired with simulated measurements.
\end{enumerate}
These three approaches are compared to the proposed method in Figure \ref{fig:diff}. Details of the experimental conditions can be found in appendix \ref{exp_1}. 
Similar experiments are then performed to reconstruct $32 \times 32$ images from the CelebA data set and the CIFAR10 data set. Different degradation conditions are tested, including blurring, down-sampling and partial occlusion. For each case, models are trained with $K = 1,000$ and $K = 10,000$ available training pairs. More details about these experiments can be found in appendix \ref{exp_1}. Reconstructions are performed with $2,000$ test examples. Figure \ref{fig:psnr} shows the average PSNR, while recovered ELBO values are reported in table \ref{tab:ELBO}.
\begin{center}
\begin{table}
\caption{Test set evidence lower bound (ELBO) for the proposed framework compared to alternative methods of using the same information to train a CVAE. The proposed framework consistently returns higher ELBO values, indicating a more accurate posterior recovery.}
\label{tab:ELBO}
\vspace{0.3cm}
  \begin{tabular}{| l | p{2.4cm} | p{2.4cm} |p{2.4cm} | p{2.4cm} |}
    \hline
     & Paired & Simulations & paired+ & Proposed  \\
     & Examples &  & Simulations &   \\
    \hline
    CelebA, K=1,000 & $-19827 \pm 771$ & $-208 \pm 403$ & $11300 \pm 454$ & $\mathbf{14553 \pm 20}$ \\
    $\times 2$ Down-sampling &  &  &  &  \\\hline
    CelebA, K=1,000 & $-21390 \pm 1655$ & $14124 \pm 33$ & $14751 \pm 42$ & $\mathbf{15134 \pm 18}$ \\
    Partial Occlusion & & & &  \\ \hline
    CelebA, K=1,000 & $-16264 \pm 122$ & $10581 \pm 21$ & $12371 \pm 532$ & $\mathbf{13365 \pm 201}$ \\
    Blurring $\sigma = 2.5px$ & & & &  \\ \hline
    CelebA, K=1,000 & $-13872 \pm 1298$ & $13152 \pm 62$ & $13805 \pm 221$ & $\mathbf{14189 \pm 63}$ \\
    Blurring $\sigma = 1.5px$ & & & &  \\ \hline
    CelebA, K=10,000 & $13450 \pm 149$ & $-208 \pm 403$ & $10303 \pm 1192$ & $\mathbf{14763 \pm 2}$ \\
    $\times 2$ Down-sampling &  &  &  &  \\\hline
    CelebA, K=10,000 & $12902 \pm 556$ & $14124 \pm 33$ & $15043 \pm 17$ & $\mathbf{15187 \pm 32}$ \\
    Partial Occlusion & & & &  \\ \hline
    CelebA, K=10,000 & $13265 \pm 53$ & $10581 \pm 21$ & $12635 \pm 437$ & $\mathbf{14672 \pm 9}$ \\
    Blurring $\sigma = 2.5px$ & & & &  \\ \hline
    CelebA, K=10,000 & $13502 \pm 310$ & $13152 \pm 62$ & $13936 \pm 136$ & $\mathbf{14842 \pm 11}$ \\
    Blurring $\sigma = 1.5px$ & & & &  \\ \hline
    CIFAR10, K=1,000 & $-21846 \pm 2128$ & $-3059 \pm 987$ & $12005 \pm 921$ & $\mathbf{14247 \pm 19}$ \\
    $\times 2$ Down-sampling &  &  &  &  \\\hline
    CIFAR10, K=1,000 & $-23358 \pm 2188$ & $12890 \pm 57$ & $14118 \pm 61$ & $\mathbf{14702 \pm 64}$ \\
    Partial Occlusion & & & &  \\ \hline
    CIFAR10, K=1,000 & $-18683 \pm 51$ & $10051 \pm 82$ & $12924 \pm 296$ & $\mathbf{13212 \pm 195}$ \\
    Blurring $\sigma = 2.5px$ & & & &  \\ \hline
    CIFAR10, K=1,000 & $-14390 \pm 40$ & $13008 \pm 105$ & $13869 \pm 174$ & $\mathbf{13988 \pm 25}$ \\
    Blurring $\sigma = 1.5px$ & & & &  \\ \hline
    CIFAR10, K=10,000 & $13496 \pm 69$ & $-3059 \pm 987$ & $12096 \pm 577$ & $\mathbf{14415 \pm 23}$ \\
    $\times 2$ Down-sampling &  &  &  &  \\\hline
    CIFAR10, K=10,000 & $12171 \pm 925$ & $12890 \pm 57$ & $14427 \pm 37$ & $\mathbf{14789 \pm 38}$ \\
    Partial Occlusion & & & &  \\ \hline
    CIFAR10, K=10,000 & $13134 \pm 219$ & $10051 \pm 82$ & $13094 \pm 312$ & $\mathbf{14348 \pm 30}$ \\
    Blurring $\sigma = 2.5px$ & & & &  \\ \hline
    CIFAR10, K=10,000 & $13402 \pm 177$ & $13008 \pm 105$ & $13974 \pm 141$ & $\mathbf{14540 \pm 21}$ \\
    Blurring $\sigma = 1.5px$ & & & &  \\ \hline
  \end{tabular}
  \vspace{-0.3cm}
  \end{table}
\end{center}

The proposed framework proved advantageous across all tested conditions, both with respect to the mean reconstruction quality, given by the mean PSNR values, and the recovered posterior density matching, approximately measured by the ELBO values. It is also noticeable how the choice of optimal approach amongst the three naive strategies is far from obvious; which training method yields best performance is highly dependent on available number $K$ of image-observation pairs and type of transformation. In contrast, the proposed framework consistently gives the best results, proving its ability to better exploit the provided information, independently of the particular conditions.

\subsection{Holographic image reconstruction}

\begin{figure}[t]
    \centering
    \includegraphics[width=0.75\linewidth]{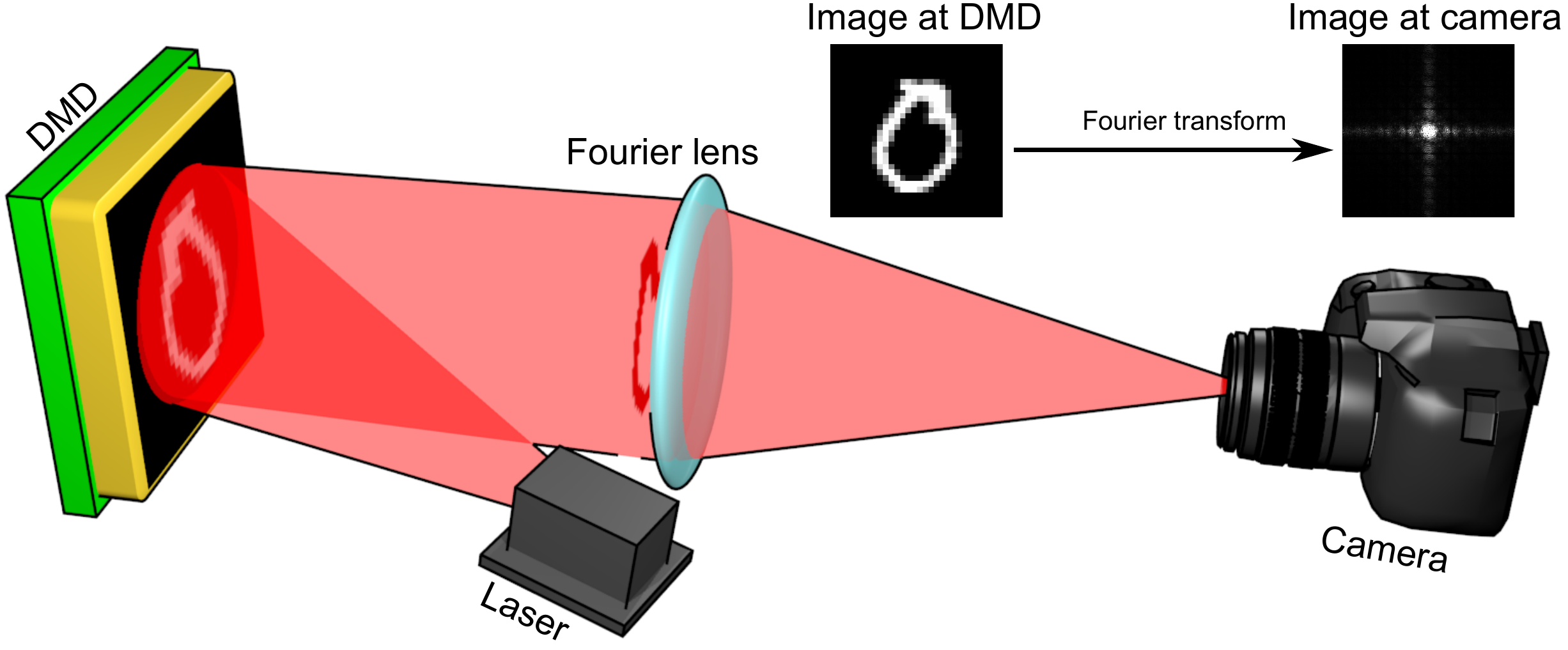}
    \caption{Experimental set up used for holographic image reconstruction. A binary amplitude image is projected by the Digital Micromirror Device (DMD) and a lens placed at the focal distance from the DMD display produces the corresponding Fourier image at the camera.}
    \label{fig:setup}
\end{figure}

Sensor arrays, such as CCD or CMOS cameras, are a ubiquitous technology that obtain a digital image of a scene. However, cameras are only able to retrieve the intensity of the light field at every point in space, computational techniques and additional elements in imaging set-ups are required to obtain the full information of the light field, i.e. both amplitude and phase. 
Unfortunately, it is not always possible to include the additional experimental components to the set-up and therefore algorithms have been adapted to use only intensity images.
Retrieving the full light field information from intensity-only measurements is a very important inverse problem that has been studied exhaustively during the last 40 years \citep{1972:optik:gs,1982:ao:fineup,2015:ieee:segev}.

Machine learning methods have been proposed in this context to learn either phase or amplitude of images/light fields from intensity-only diffraction patterns recorded with a camera  \citep{2017:optica:barbastathis,2018:lsa:ozcan,2019:lsa:ozcan}. Such an ability is desirable because the intensity images can be recorded with cheap digital cameras, instead of expensive and delicate phase-sensitive instruments. Following these recent advances, we aim to use our proposed variational framework approach to solve the following problem: 
Given the camera intensity image of the diffraction pattern at the Fourier plane, what is the amplitude of the corresponding projected image? This apparently simple problem has multiple applications in areas such as material science, where X-rays are used to infer the structure of a molecule from its diffraction pattern \citep{2003:oe:xray,2019:candes}, optical trapping \citep{2008:oe:ritsch-marte}, and microscopy \citep{2004:prl:faulker}. 

\subsubsection{Experimental set-up and Data}

The experiment consisted of an expanded laser beam incident onto a Digital Micromirror Device (DMD) which displays binary patterns, as shown in Figure~\ref{fig:setup}. DMDs consist of an array of micron-sized mirrors that can be arranged into two angles that correspond to ``on" and ``off" states of the micromirror.
Consequently, the amplitude of the light is binarised by the DMD pattern and propagates toward a single lens. The lens, placed at the focal distance from the DMD display, will cause the rays to form the Fourier image of the MNIST digit at the camera. 

\begin{figure*}[t]
  \centering
  \includegraphics[width=\linewidth]{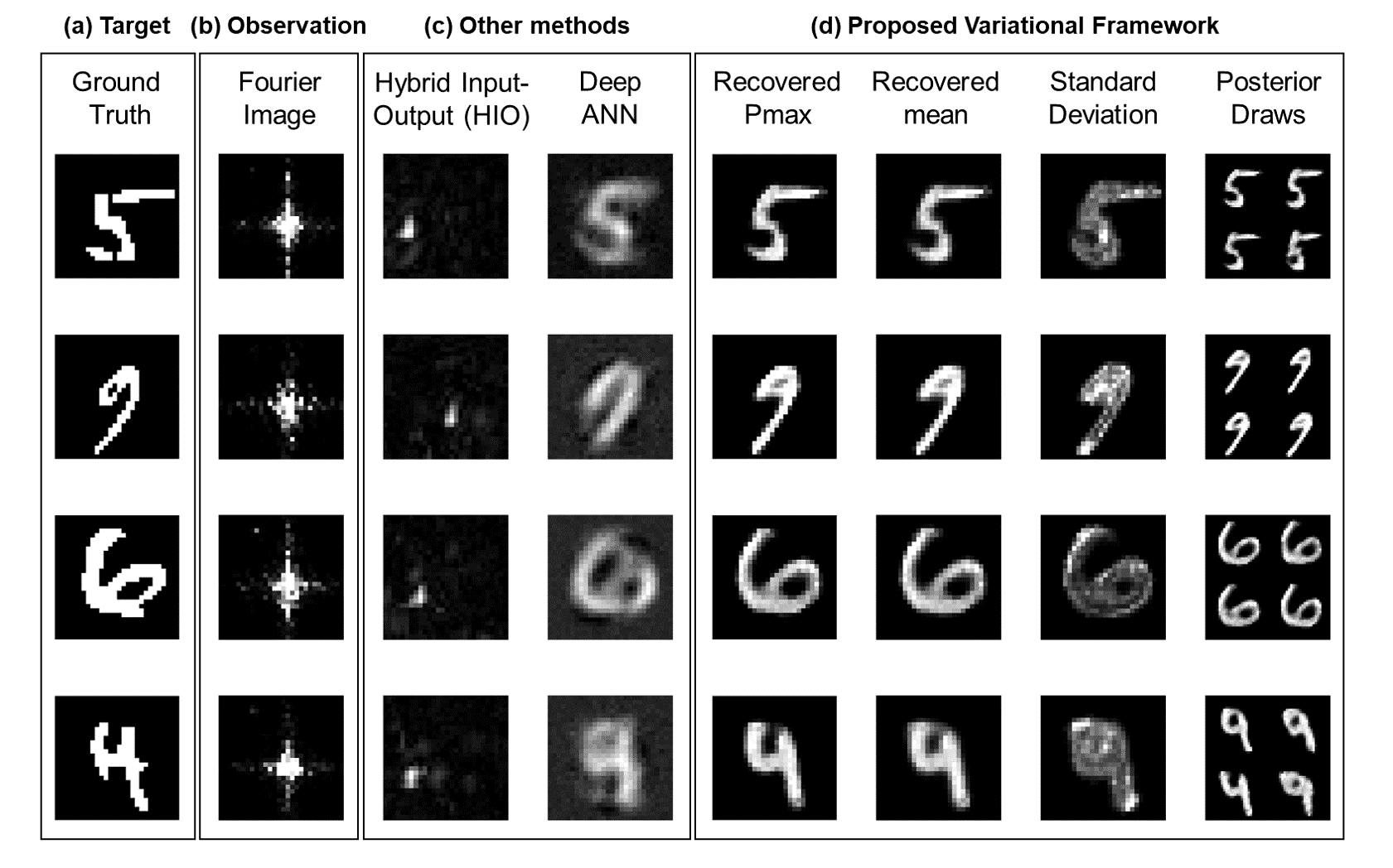}
  \caption{Reconstructions from experimental Fourier intensity image data. (a) Target images projected by the DMD, (b) intensity Fourier image observed at the camera, (c)  reconstructions using other techniques (d) reconstructions obtained with the proposed variational method, including pseudo-maximum, pixel-marginal mean, pixel-marginal standard deviation and examples of draws from the recovered posterior.}
\label{fig:FFT_main}
\end{figure*}

To make the problem even harder, we work under saturation conditions, i.e. assuming blinding on the camera, and with extremely low-resolution images. 
We display $9600$ MNIST digits on the DMD and record the corresponding camera observations. This data is used as high fidelity paired ground truths $X^*$ and measurements $Y^*$. The remaining $50400$ MNIST examples are used as the large set of unobserved ground truth signals $X$. The analytical observation model $p(\widetilde{y}|x)$ is built as a simple intensity Fourier transform computation, to which we add artificial saturation.

\subsubsection{Reconstruction}

\begin{figure*}[t]
  \centering
  \includegraphics[width=\linewidth]{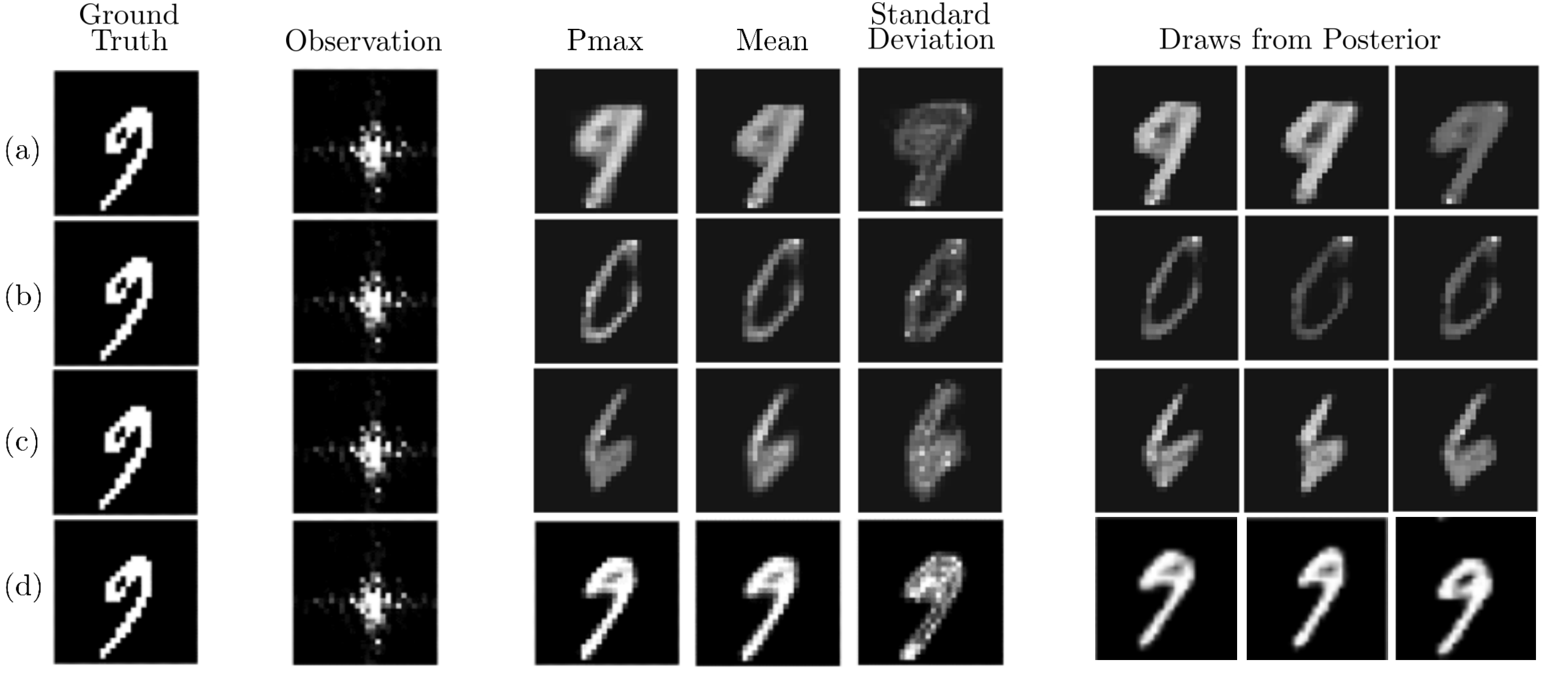}
  \caption{Image posterior recovery from phase-less measurements. (a) CVAE trained with the available $K = 9,600$ paired examples alone. The size of this training set is too small to obtain accurate posteriors. (b) CVAE trained with $L = 50,400$ target examples and corresponding simulated observations from the inaccurate observation model. Because the observation model, i.e. a simple Fourier transform, does not match the true one encountered upon testing, the image is not well recovered. (c) CVAE trained with $K = 9,600$ paired examples in combination with $L = 50,600$ target examples and corresponding inaccurately simulated observations. The presence of real measurements in the training gives more realistic MNIST-like shapes, but the reconstruction is still inaccurate. (d) CVAE trained with proposed variational framework. The sources of information are exploited in a principled way, resulting in accurate posterior recovery.}
\label{fig:holo}
\end{figure*}

Image reconstruction from the image at the Fourier plane is performed with the proposed variational framework and compared with a Hybrid Input-Output (HIO) complex light-field retrieval algorithm and a 4-layer deep Artificial Neural Network (deep ANN) as shown in Figure~\ref{fig:FFT_main}.
On the one hand, given that the HIO retrieval algorithm is an iterative method that uses the light intensity pattern recorded by the camera at the Fourier plane at each iteration, it is not expected to operate well in conditions of saturation and/or down-sampling (see appendix \ref{exp_2} for details). This is precisely what we observe in Figure~\ref{fig:FFT_main}(c), where the HIO algorithm simply predicts spots at some positions.

The results of a deep ANN show that a more accurate solution can be found. However, the accuracy of the deep ANN to reconstruct ground truth is hindered by the limited training set of 9600 experimental images. As shown in Figure \ref{fig:FFT_main}(d), highly accurate reconstructed images are achieved with the proposed variational method which exploits the generative multi-fidelity forward model to train the inverse model using the additional unobserved 50400 examples. Furthermore, the proposed method retrieves full posterior densities, from which we can draw to explore different possible reconstructions as a result of the ill-posed nature of the inverse problem. 

In order to demonstrate the advantage of employing the proposed framework compared to naive strategies in a real scenario, we repeat the evaluation of figure \ref{fig:diff} for this physical experiment. An example is shown in figure \ref{fig:holo}. Analogously to the simulated experiments, using the experimental training set alone gives results of limited quality. Combining the simulations and real data in naive ways completely disrupts reconstructions, as in this experiments the simulations are significantly different from real measurements. However, they are far from useless, as including them in a principled way through the proposed framework gives significant improvement in reconstruction quality.

\begin{figure*}[t]
  \centering
  \includegraphics[width=\linewidth]{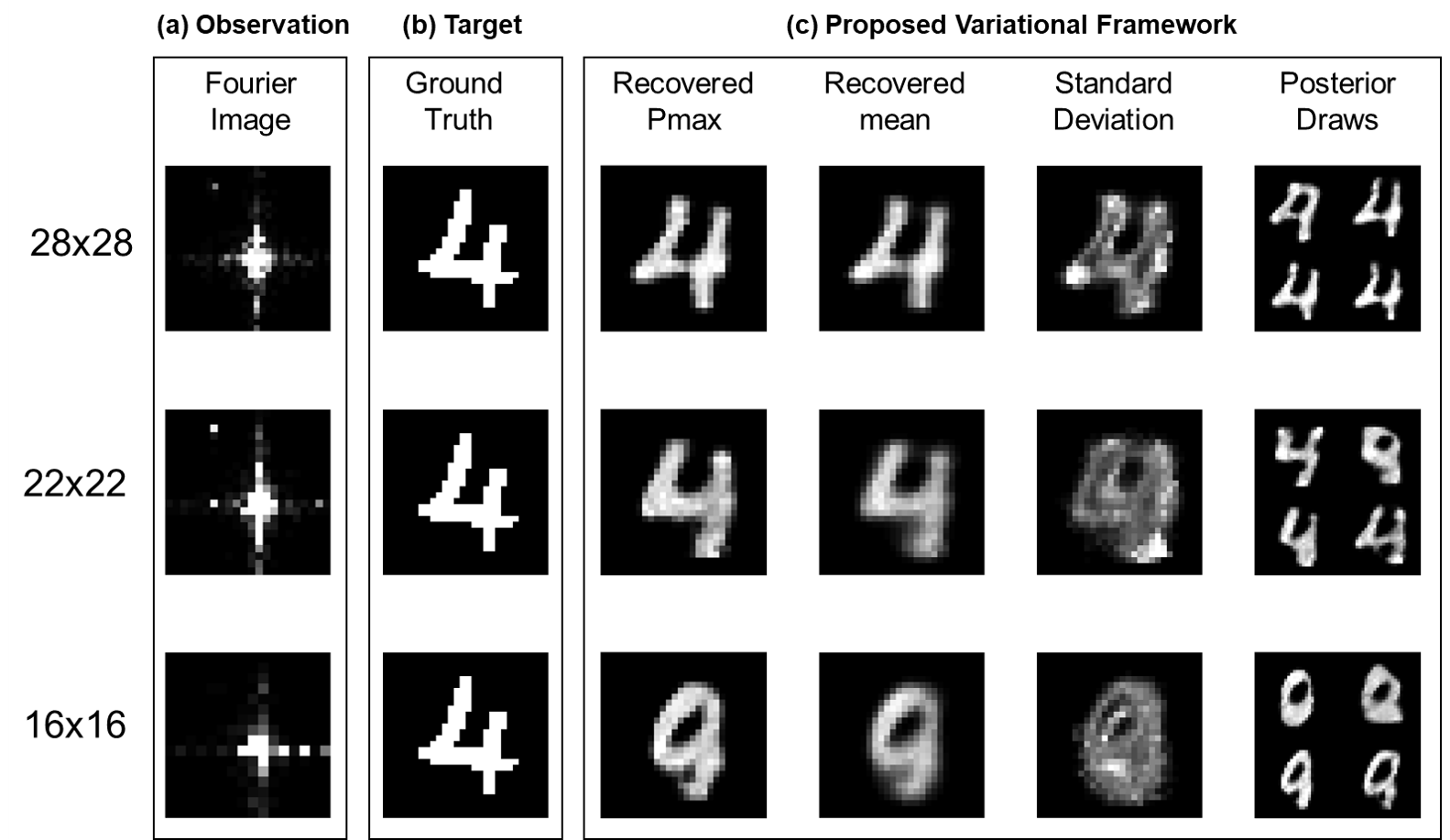}
  \caption{ (a) Experimental Fourier intensity image data down-sampled to $(28\times28)$, $(22\times22)$ and $(16\times16)$ (top to bottom) for (b) the same target image. (c) The proposed variational framework which shows the reconstructed image quality degrades with decreasing resolution of the measured data. As expected, the standard deviation and samples from the recovered posterior show high variability to the solution when reaching the critically ill-posed resolution limit of $(16\times16)$.}
\label{fig:FFT_res}
\end{figure*}

Figure \ref{fig:FFT_res} illustrates further this posterior exploration capability. When progressively down-sampling the resolution of experimentally measured observations, the pseudo-max reconstructed image quality degrades and the range of possible solutions, visualised through the different draws, extends. When down-sampling the experimental images to a resolution of $(16 \times 16)$, the inverse problem becomes critically ill-posed such that the solution space becomes too varied to accurately recover the ground truth image.

\subsection{Imaging Through Highly Scattering Media}

Imaging through strongly diffusive media remains an outstanding problem in optical CI, with applications in biological and medical imaging and imaging in adverse environmental conditions \citep{DOT_BOOK}. Visible or near-infrared light does propagate in turbid media, such as biological tissue or fog, however, its path is strongly affected by scattering, leading to the loss of any direct image information after a short propagation length. The reconstruction of a hidden object from observations at the scattering medium's surface is the inverse problem that will be addressed in this section.

\subsubsection{Physical Experiment}

\begin{figure*}[t]
  \centering
  \includegraphics[width=\linewidth]{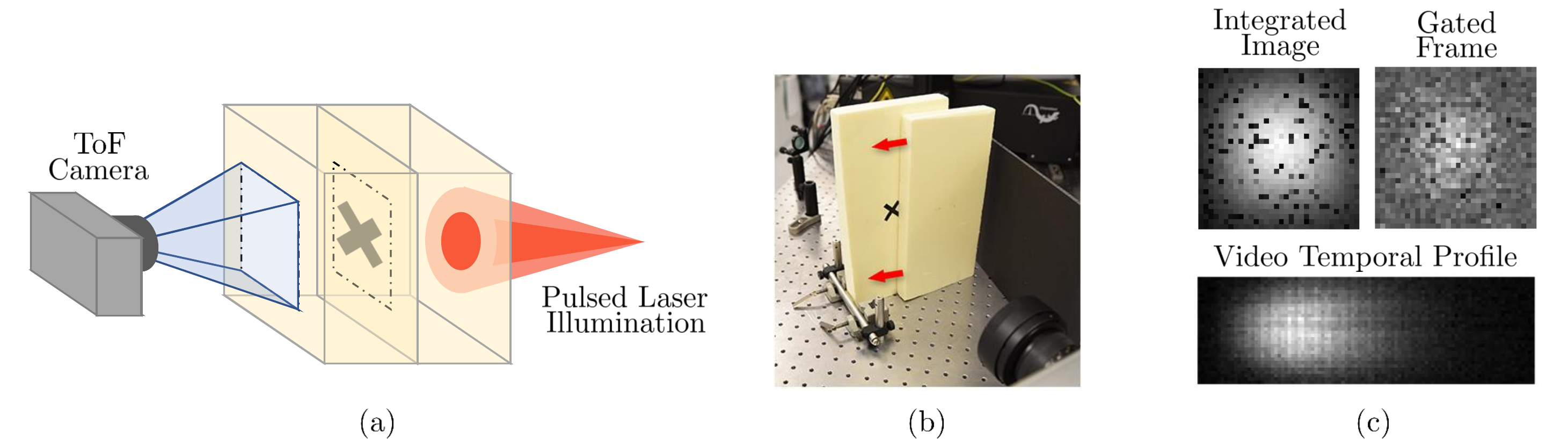}
\caption{Experimental set up for imaging through scattering media. (a) Schematic representation of the experiment. A target object is embedded between two \unit{2.5}{\centi\metre}-thick slabs of diffusing material, with absorption and scattering properties comparable to those of biological tissue. One exposed face is illuminated with a pulsed laser and the opposite face is imaged with the ToF camera. (b) A photograph of the same experimental set up. (c) Example of the video recorded by the ToF camera as light exits the medium's surface. Images show the integration over all time frames (i.e. the image a camera with no temporal resolution would acquire), a single frame of the video gated in time and the intensity profile of a pixels' line at different times.}
\label{fig:CDI_exp}
\end{figure*}

Following the experimental implementation presented by \citet{DOT_NAT}, imaging is performed with a \unit{130}{\femto\second} near-infrared pulsed laser and a single photon sensitive time of flight (ToF) camera with a temporal resolution of \unit{55}{\pico\second} to perform transmission diffuse imaging. In these experiments, different cut-out shapes of alphabetic letters were placed between two identical \unit{2.5}{\centi\metre} thick slabs of diffusive material, with measured absorption and scattering coefficients of $\mu_a = $~\unit{0.09}{\centi\reciprocal\metre} and $\mu_s = $~\unit{16.5}{\centi\reciprocal\metre} respectively. A schematic representation and a photograph of the set up are shown in Figure \ref{fig:CDI_exp}(a-b).

A pulse of light from the laser propagates through the diffusing material, reaches the hidden object, which partially absorbs it, and then propagates further through the medium to the imaged surface. The ToF camera records a video of the light intensity as a function of time as it exits the medium. A video recorded with an empty piece of material is used as background and subtracted to that obtained with the object present, thereby obtaining a video of the estimated difference in light intensity caused by the hidden object. An example of such videos is shown in Figure \ref{fig:CDI_exp}(c). At this depth, more than $40$ times longer than the photon's mean free path, the diffusion effect is so severe that even basic shapes are not distinguishable directly from the videos. Furthermore, the measurements experience low signal-to-noise ratio due to the low light intensity that reaches the imaged surface and the low fill factor of the ToF camera, which is about $1 \%$. Achieving accurate reconstructions with simple objects in this settings, is a first important step towards achieving imaging through biological tissue with near-infrared light and hence non-ionising radiation.

\subsubsection{Training Data and Models}

\begin{figure*}[h]
  \centering
  \includegraphics[width=\linewidth]{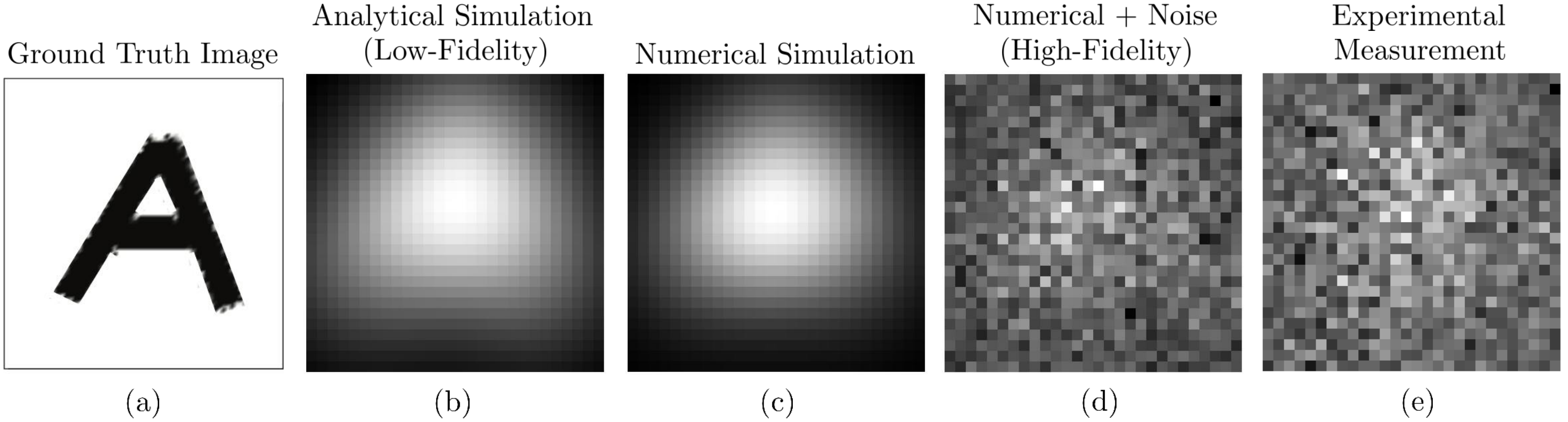}
\caption{Simulated and real measurements from the time of flight (ToF) Camera. Images are single frames from the camera videos. (a) Image of the hidden object. (b) Simulated measurement using the analytical solution from the linear approximation, taken as low-fidelity estimate. (c) Simulation obtained by numerically propagating the diffusion equation, which is accurate, but expensive. (d) Numerical simulation with added noise, used as high-fidelity estimates of the measurements. (e) The real measurements recorded by the ToF camera for this object.}
\label{fig:CDI_sim}
\end{figure*}

As target objects in these experiments are character-like shapes, the training images are taken from the NIST data set of hand-written characters \citep{NIST}. $86,400$ NIST images are used as the large data set of unobserved target examples $X$. Because of experimental preparation, it is infeasible to perform a large number of physical acquisitions to build a training set. However, the process of light propagation through a highly scattering medium can be accurately described with the diffusion approximation, commonly adopted in these settings \citep{DOT_NAT,diff_app}. The propagation of photons under this assumption is described by the following differential equation
\begin{equation}
	\label{diff_eq}
	c^{-1}\frac{\partial \Phi(\vec{r},t)}{\partial t}+\mu_{a}\Phi(\vec{r},t)- D\nabla \cdot \left[\nabla\Phi(\vec{r},t)\right]=S(\vec{r},t),
\end{equation}
where $c$ is the speed of light in the medium, $\vec{r}$ is the spatial position, $t$ is the temporal coordinate, $\Phi(\vec{r},t)$ is the photons flux, $S(\vec{r},t)$ is a photon source, here the illumination at the surface, and $ D = \big( 3 (\mu_a + \mu_s) \big)^{-1} $. The measurements recorded by the ToF camera in the experiment described above can be accurately simulated by numerically propagating the photon flux $\Phi(\vec{r},t)$ in space and time with appropriate boundary conditions at the edges of the medium and a high absorption coefficient $\mu_a$ assigned to the object voxels. These simulations are accurate, but expensive. To simulate the experiments of interest here they take in the order of a few minutes per example to run on a TitanX GPU. Obtaining paired inputs and outputs for tens of thousands of experiments is expensive. Instead, only $1,000$ examples of the $84,400$ training targets were generated in this way and were taken as high-fidelity measurement estimates $Y^*$ from corresponding ground truth images $X^*$. An example of such simulations for one of the test characters is shown in Figure \ref{fig:CDI_sim}(c-d).

In order to simulate measurements at a lower computational cost, a linear approximation of the observation process can be exploited \citep{DOT_NAT,diff_app}. For a delta function initial illumination $S(\vec{r},t) = \delta(\vec{r} = \vec{r}',t = t')$ and an infinite uniform scattering medium, an analytical solution for $\Phi(\vec{r},t)$ exists:
\begin{equation}
	\label{an_sol}
	\Phi(\vec{r},t;\vec{r}',t')= \frac{c}{\left[4\pi Dc(t-t')\right]^{3/2}} \times  
	\exp\left[-\frac{\lvert\vec{r}-\vec{r}'\rvert ^{2}}{4Dc(t-t')}\right]\exp\left[-\mu_{a}c(t-t')\right].
\end{equation}
This solution constitutes a point spread function with which an analytical estimate of the measurements can be computed through two consecutive convolutions. First, the illumination at the entering surface is convolved in 2D and time with the PSF of equation \ref{an_sol} to obtain an estimate of the illumination at the object plane. Second, this estimate multiplied by the object image at each time frame is convolved again with the PSF to estimate the intensity field at the exiting surface, imaged by the ToF camera \citep{DOT_NAT}. An example of such analytical estimates of the measurements is shown if Figure \ref{fig:CDI_sim}(b). These computations are much less expensive to perform than propagating numerically the diffusion equation, requiring less than \unit{100}{\milli\second} per sample to run on a TitanX GPU. However, they introduce approximations which sacrifice the accuracy of the simulated measurements. In particular, they don't take into account any boundary condition and assume that the observation process is linear, whereas in reality the light absorbed by some part of the object will affect the illumination at some other part. This analytical observation model is taken as the approximate likelihood $p(\widetilde{y}|x)$ generating low-fidelity measurement's estimates $\widetilde{y}$. 

\subsubsection{Results}

\begin{figure*}[h]
  \centering
  \includegraphics[width=\linewidth]{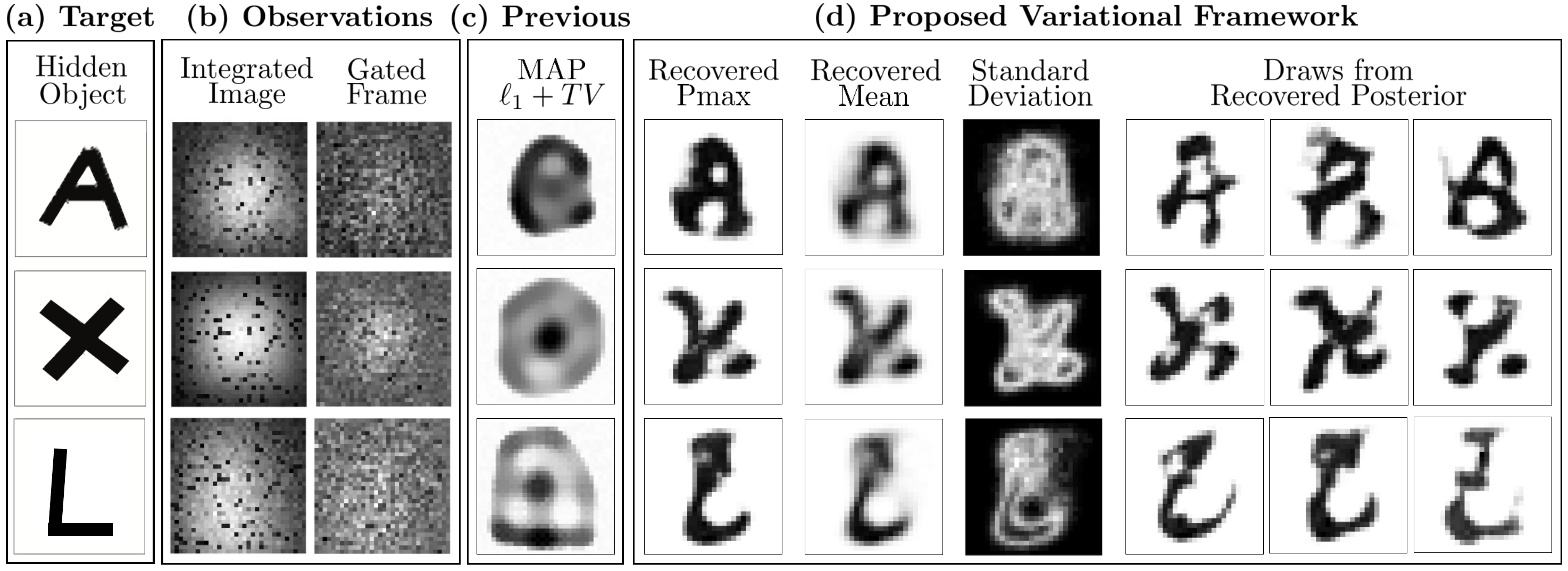}
\caption{Reconstructions from experimental ToF videos. (a) Target images embedded in the scattering medium, (b) integrated and gated frames from the ToF camera videos, constituting the observed measurements, (c) reconstruction obtained using constrained optimisation with $\ell_1$-norm and total variation regularisation and (d) reconstructions obtained with the proposed variational method, including pseudo-maximum, pixel-marginal mean, pixel-marginal standard deviation and examples of draws from the recovered posterior. The proposed framework recovers arguably more accurate images compared to the state of the art, while also allowing exploration of the manifold of possible solutions to the inverse problem.}
\label{fig:CDI_results}
\end{figure*}

The ToF videos recorded for three different shapes embedded in the scattering medium were used to perform reconstructions. Firstly, the recovery is performed using the method presented by \cite{DOT_NAT}, consisting of a constrained minimisation with $\ell_1$ and total variation regularisation. Secondly, retrieval is performed with a CVAE trained with the proposed framework and using the sources of information described above. Results are shown in Figure \ref{fig:CDI_results}.

The prior method is capable of retrieving general features of the objects embedded in the scattering medium, but sometimes results in severe artefacts that make the images unrecognisable. Furthermore, to obtain the displayed results, it is necessary to carefully tune the penalty coefficients of the constrained optimisation for each example, making such retrieval highly dependent on human supervision. Exploiting a more specific empirical prior, the proposed framework allows to retrieve more accurate reconstructions, where the different letters are clearly recognisable. Moreover, this particularly ill-posed inverse problem example highlights the importance of using a Bayesian approach; the solution space given a diffuse ToF video is rather variable and, unlike constrained optimisation and other single estimate methods, through the approximate posterior such variability can be captured by empirically estimating uncertainty and visualising different drawn samples, as shown in Figure~\ref{fig:CDI_results}(d). Note that, thanks to the proposed framework, the model was successfully trained with very limited effort and resources; the large data set of targets was readily available independently of the application of interest, while only $1,000$ expensive simulations were used, requiring just a few tens of hours of simulation time on a single GPU to be generated. 

\section{Conclusion}

This paper introduced a novel framework to train variational inference in imaging inverse problems, utilising different types of data and domain expertise in combination. As a result, Bayesian machine learning is rendered accessible for a broad range of imaging applications, where empirical training data is typically scarce or expensive to collect. The method was derived from a Bayesian formulation of inverse problems and interpreting accessible sources of information as approximations to or samples from underlying distributions, providing theoretical foundation. Simulated experiments thoroughly tested the proposed technique in a range of different conditions, proving its ability to better exploit all sources of information available.
The method was then applied to real imaging systems, demonstrating the first successful application of Bayesian machine learning in both phase-less holographic image reconstruction and imaging through scattering media. In both settings, state of the art reconstructions were achieved, while requiring little training collection efforts, whereas before Bayesian methods would have required prohibitively large volumes of data.

\section*{Acknowledgements}
We would like to thank the editor and reviewers for their thoughtful, constructive and detailed reviews which have improved the final paper. F.T., R.M-S., D.F.  acknowledge  funding  from  Amazon  and  EPSRC  grants EP/M01326X/1, EP/T00097X/1 ({\it QuantIC}, the {\it UK Quantum Technology Hub in Quantum Enhanced Imaging}) and EP/R018634/1 ({\it Closed-Loop Data Science for Complex, Computationally- and Data-Intensive Analytics}). D.F. is supported by the Royal Academy of Engineering under the {\it Chairs in Emerging Technologies} scheme.  J.R. is supported by the EPSRC {\it CDT in Intelligent Sensing and Measurement}, Grant Number EP/L016753/1. A.T. is supported by a Lord Kelvin Adam Smith Fellowship from the University of Glasgow.

\newpage
\appendix

{\Large\bf\centering Supplementary Material \par}


\section{Details of ELBO Formulation}

\subsection{VAE Formulation for Multi-Fidelity Forward Model} \label{VAE_for}

Through Jensen's inequality, a lower bound for the parametric distribution $p_{\alpha}(y|x)$ can be defined as
\begin{eqnarray}\label{fm4}
\begin{split}
 \log p_{\alpha}(y_k|x_k) &\geq \int p(\widetilde{y}|x_k)\int q_{\beta}(w|x_k,y_k,\widetilde{y}) \log \left[ \frac{p_{\alpha_1}(w|x_k,\widetilde{y})}{q_{\beta}(w|x_k,y_k,\widetilde{y})}p_{\alpha_2}(y_k|x_k,\widetilde{y},w) \right]dw d \widetilde{y} \\
 &= \int p(\widetilde{y}|x_k) \left[\int  q_{\beta}(w|x_k,y_k,\widetilde{y}) \log p_{\alpha_2}(y_k|x_k,\widetilde{y},w)dw - D_{KL}(q_{\beta} || p_{\alpha_1}) \right] d\widetilde{y},
 \end{split}
\end{eqnarray}
where $q_{\beta}(w|x,y,\widetilde{y})$ is the recognition model, chosen as an isotropic Gaussian distribution, the moments of which are outputs of a neural network taking as inputs targets $x$, high-fidelity measurements $y$ and low-fidelity measurements $\widetilde{y}$. $D_{KL}(q_{\beta} || p_{\alpha_1})$ is the KL divergence between the distributions $q_{\beta}$ and $p_{\alpha_1}$ defined as 
\begin{equation}\label{fm5}
	D_{KL}(q_{\beta} || p_{\alpha_1}) = \int q_{\beta}(w|x_k,y_k,\widetilde{y}) \log \frac{q_{\beta}(w|x_k,y_k,\widetilde{y})}{p_{\alpha_1}(w|x_k,\widetilde{y})}dw.
\end{equation}
As both $p_{\alpha_1}(w|x_k,\widetilde{y})$ and $q_{\beta}(w|x_k,y_k,\widetilde{y})$ are isotropic Gaussian distributions, a closed form solution for the KL divergence exists and can be exploited in computing and optimising the lower bound \citep{VAE}.

\subsection{VAE Formulation for Variational Inverse Model} \label{VAE_inv}

using Jensen's inequality, a tractable lower bound for the expression of equation \ref{max_lik} can be derived with the aid of a parametric recognition model $q_{\phi}(z|x,y)$ as

\begin{eqnarray}\label{VICI_ELBO}
\begin{split}
  \int p(x) \int p_{\alpha}(y|x) \log \int r_{\theta_1}(z|y) r_{\theta_2}(x|z,y) dz dy dx & \geq \\ \int p(x) \int p_{\alpha}(y|x) \int q_{\phi}(z|x,y) \log \left[ \frac{r_{\theta_1}(z|y)}{q_{\phi}(z|x,y)} r_{\theta_2}(x|z,y) \right] dz dy dx & = \\
  \int p(x) \int p_{\alpha}(y|x)  \left[\int q_{\phi}(z|x,y) \log r_{\theta_2}(x|z,y) dz - D_{KL}(q_{\phi}(z|x,y) || r_{\theta_1}(z|y)) \right] dy dx. & 
\end{split}
\end{eqnarray}
The recognition model $q_{\phi}(z|x,y)$ is an isotropic Gaussian distribution in the latent space, with moments inferred by a neural network, taking as input both example targets $x$ and corresponding observations $y$. This neural network may be fully connected, partly convolutional or completely convolutional, depending on the nature of the targets $x$ and observations $y$. As both $q_{\phi}(z|x,y)$ and $r_{\theta_1}(z|y)$ are isotropic Gaussians in the latent space, their KL divergence $D_{KL}(q_{\phi}(z|x,y)$ has an analytical solution. All the remaining integrals can be estimated stochastically, leading to the maximisation of equation \ref{VICI_obj}.

\section{Algorithms}\label{alg}

In this supplementary section we detail the training procedure for the forward model $p_{\alpha}(y|x)$ and inverse model $r_{\theta}(x|y)$. The following pseudo-code details the training of the two models:

\begin{algorithm}[h!]
\caption{Training the Forward Model $p_{\alpha}(y|x)$}\label{FM_alg}

\vspace{0.2cm}

\textbf{\textit{Inputs:}} Analytical forward model from domain expertise $p(\widetilde{y}|x)$; set of measured Ground-truths $X^* = \{ x_{k=1:K} \}$; corresponding set of measurements $Y^* = \{ y_{k=1:K} \}$; user-defined number of iterations $N_{iter}$; batch zise $K_{b} \leq K$; Initialised weights  \{$\alpha_1^{(0)}, \alpha_2^{(0)}, \beta^{(0)}$\}; user-defined latent dimensionality, $J_w$.
\vspace{0.2cm}
%
\begin{algorithmic}[1]
%
\State \textbf{for} \textit{the} $n$'th \textit{iteration} \textbf{in} $[0:N_{iter}]$ \\
\quad \textbf{for} \textit{the} $k$'th \textit{example} \textbf{in} $[0:K_{b}]$ \\
\quad \quad $\widetilde{y}_{k} \sim p(\widetilde{y}|x_k)$ \\
\quad \quad \textit{compute moments of} $p_{\alpha_1^{(n)}}(w|x_k,\widetilde{y}_{k})$ \\
\quad \quad \textit{compute moments of} $q_{\beta^{(n)}}(w|x_k, y_k,\widetilde{y}_{k})$ \\
\quad \quad $w_{k} \sim q_{\beta^{(n)}}(w|x_k, y_k,\widetilde{y}_{k})$ \\
\quad \quad \textit{compute moments of} $p_{\alpha_2^{(n)}}(y|x_k,\widetilde{y}_{k},w_k)$ \\
\quad \textbf{end} \\
\quad $\textbf{L}^{(n)} \gets \frac{1}{K_{b}} \sum_k^{K_{b}}  \log p_{\alpha_2^{(n)}}(y|x_k,\widetilde{y}_{k},w_k) - D_{KL}(q_{\beta^{(n)}}(w|x_k, y_k,\widetilde{y}_{k})|| p_{\alpha_1^{(n)}}(w|x_k,\widetilde{y}_{k}))$  \\
\quad $\alpha_1^{(n+1)}, \alpha_2^{(n+1)}, \beta^{(n+1)} \gets {\arg\max} (\textbf{L}^{(n)})$ \\
\textbf{end}
\end{algorithmic}
\end{algorithm}

\begin{algorithm}[h!]
\caption{Training the Inverse Model $r_{\theta}(x|y)$}\label{IM_alg}

\vspace{0.2cm}

\textbf{\textit{Inputs:}} Trained multi-fidelity forward model $p_{\alpha}(y|x)$; set of unobserved ground-truths $X = \{ x_{l=1:L} \}$; user-defined number of iterations $N_{iter}$; batch zise $L_{b} \leq L$; Initialised weights  \{$\theta_1^{(0)}, \theta_2^{(0)}, \phi^{(0)}$\}; user-defined latent dimensionality, $J_z$.
\vspace{0.2cm}
%
\begin{algorithmic}[1]
%
\State \textbf{for} \textit{the} $n$'th \textit{iteration} \textbf{in} $[0:N_{iter}]$ \\
\quad \textbf{for} \textit{the} $k$'th \textit{example} \textbf{in} $[0:K_{b}]$ \\
\quad \quad $y_l \sim p_{\alpha}(y|x_l)$ \\
\quad \quad \textit{compute moments of} $r_{\theta_1^{(n)}}(z|y_l)$ \\
\quad \quad \textit{compute moments of} $q_{\phi^{(n)}}(z|x_l, y_l)$ \\
\quad \quad $z_{l} \sim q_{\phi^{(n)}}(z|x_l, y_l)$ \\
\quad \quad \textit{compute moments of} $r_{\theta_2^{(n)}}(x|z_l,y_l)$ \\
\quad \textbf{end} \\

\quad $\textbf{L}^{(n)} \gets \frac{1}{L_{b}} \sum_l^{L_{b}}  \log r_{\theta_2^{(n)}}(x|z_l,y_l) - D_{KL}(q_{\phi^{(n)}}(z|x_l, y_l)|| r_{\theta_1^{(n)}}(z|y_l))$  \\
\quad $\theta_1^{(n+1)}, \theta_2^{(n+1)}, \phi^{(n+1)} \gets {\arg\max} (\textbf{L}^{(n)})$ \\
\textbf{end}
\end{algorithmic}
\end{algorithm}

\section{Details of the Models' Architectures}

The different architectures for each inference distribution implemented in the presented experiments are described here.

\subsection{Multi-Fidelity Forward Model}

\begin{figure}[h]
  \centering
  \includegraphics[width=\linewidth]{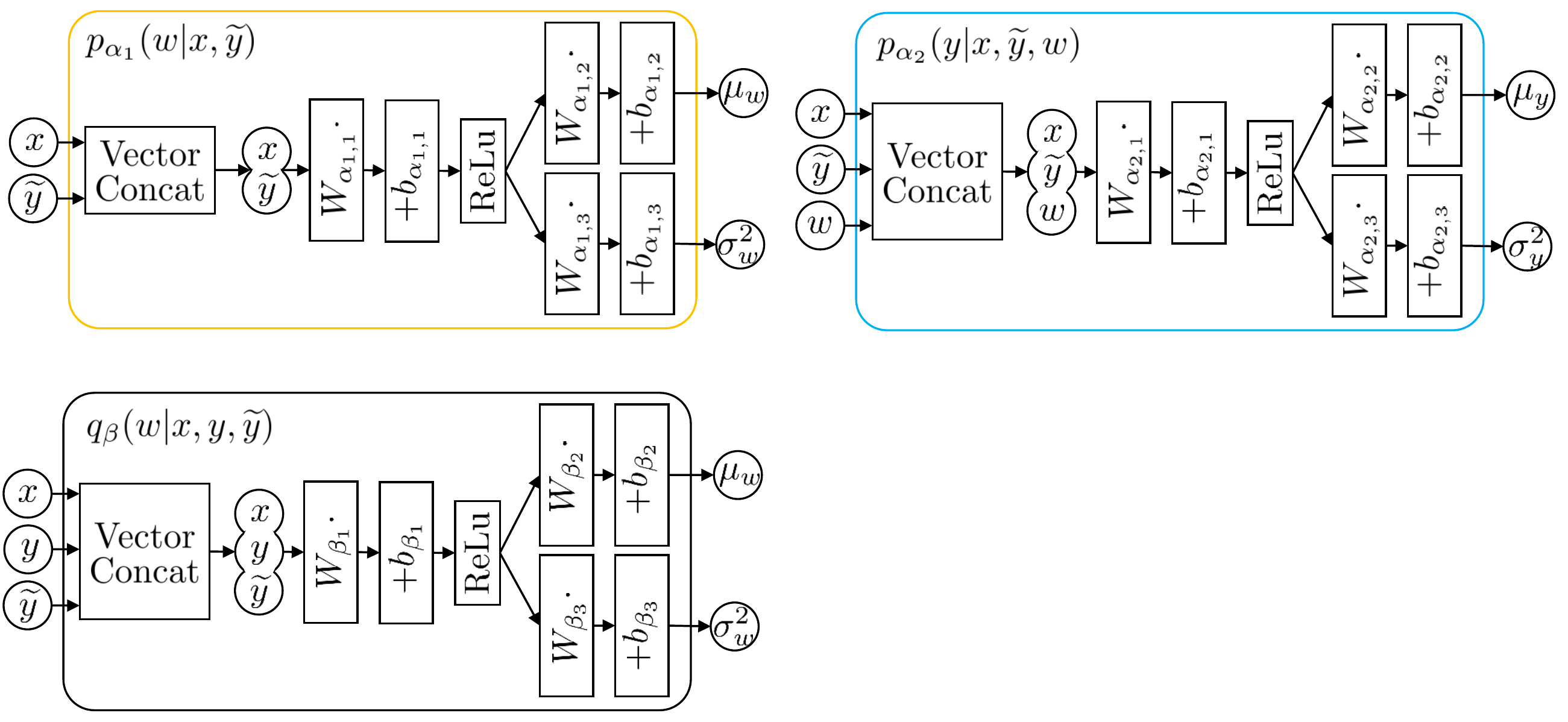}
\caption{Parametric distributions' structures for the fully connected version of the multi-fidelity forward model. The output variables are sampled from Gaussian distributions having the corresponding output moments shown.}
\label{fig:supp_forward_fullyconn}
\end{figure}

\begin{figure}[h]
  \centering
  \includegraphics[width=10cm]{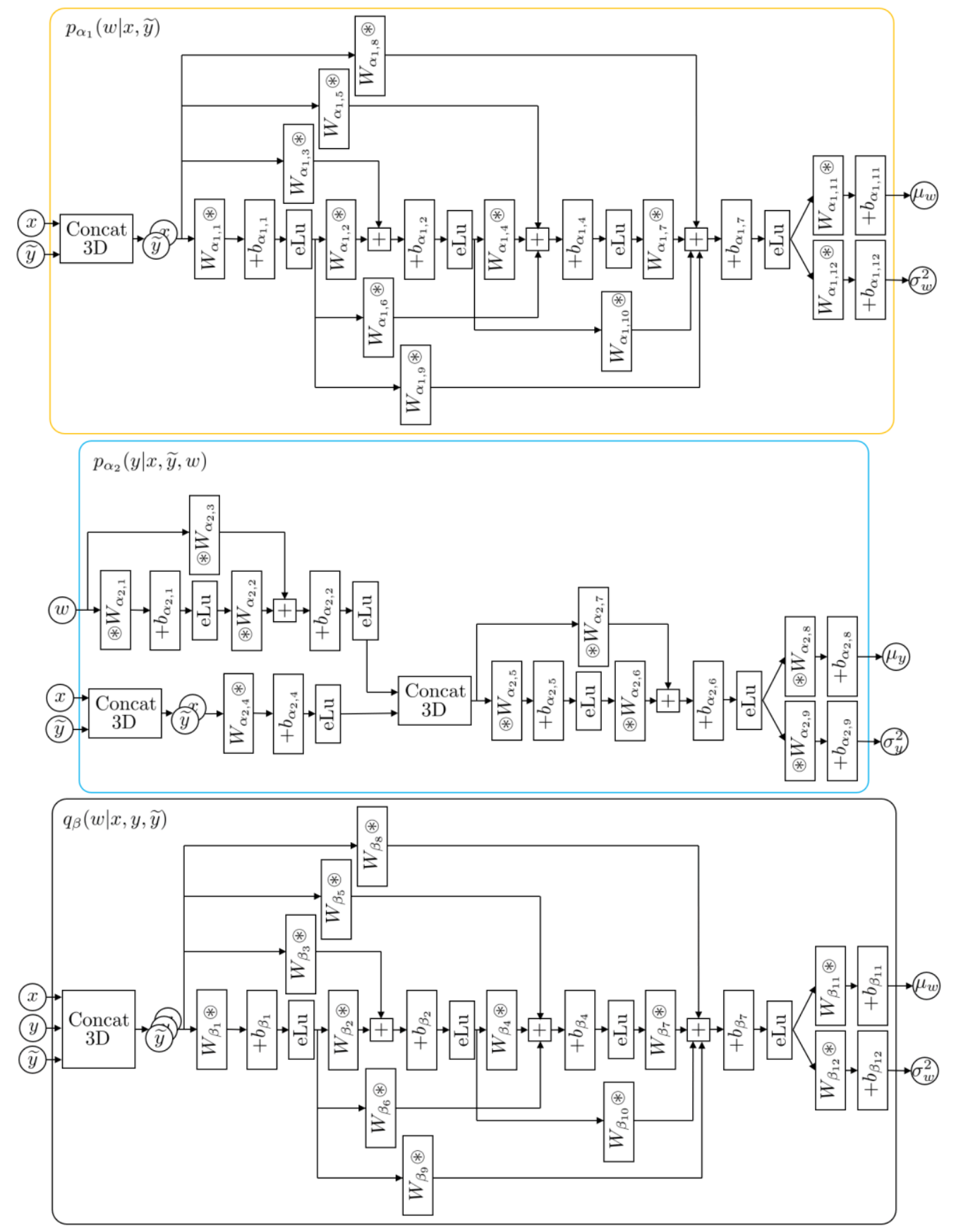}
\caption{Parametric distributions' structures for the convolutional version of the multi-fidelity forward model. $W \circledast$ indicates a convolution with filter bank $W$, while $\circledast W$ indicates a transpose convolution.}
\label{fig:supp_forward_conv}
\end{figure}

The multi-fidelity forward model includes three parametric distributions, the parameters of which are optimised during training (see figure \ref{fig:multifidelity}); $p_{\alpha_1}(w|x,\widetilde{y})$,  $p_{\alpha_2}(y|x,\widetilde{y},w)$ and $q_{\beta}(w|x,y,\widetilde{y})$. Two versions of the multi-fidelity forward model were implemented. In the first, the parametric distributions consist of fully connected layers mapping inputs to outputs' Gaussian moments, from which samples are drawn upon training and inference. These structures are schematically represented in figure \ref{fig:supp_forward_fullyconn}. In the second, the parametric distributions consist of deeper convolutional recurrent layers, again mapping mapping inputs to outputs' Gaussian moments, from which samples are drawn upon training and inference. These structures are instead shown in figure \ref{fig:supp_forward_conv}.

\subsection{Variational Inverse Model}

\begin{figure*}[h]
  \centering
  \includegraphics[width=\linewidth]{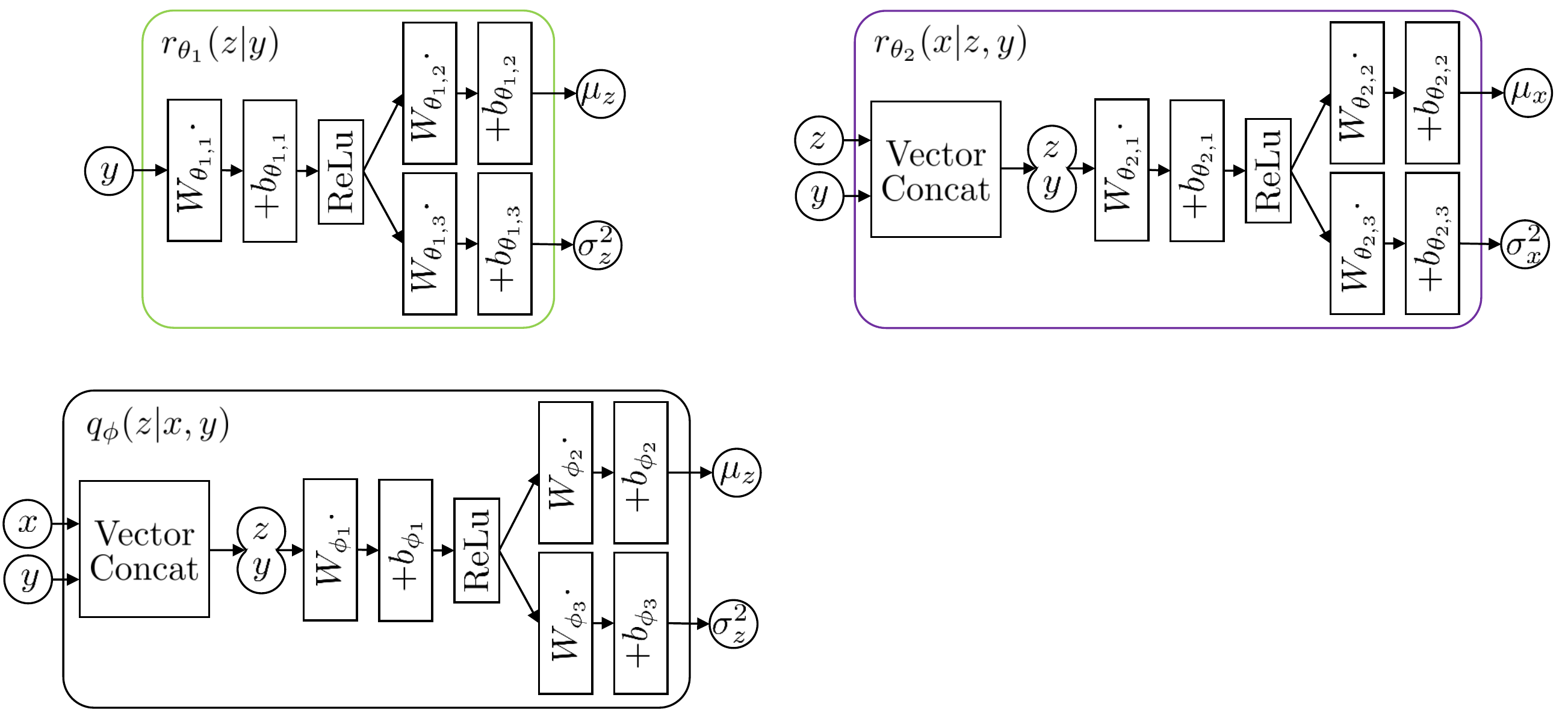}
\caption{Parametric distributions' structures for the fully connected version of the inverse model. The output variables are sampled from Gaussian distributions having the corresponding output moments shown.}
\label{fig:supp_inverse_fullyconn}
\end{figure*}

\begin{figure*}[h!]
  \centering
  \includegraphics[width=10cm]{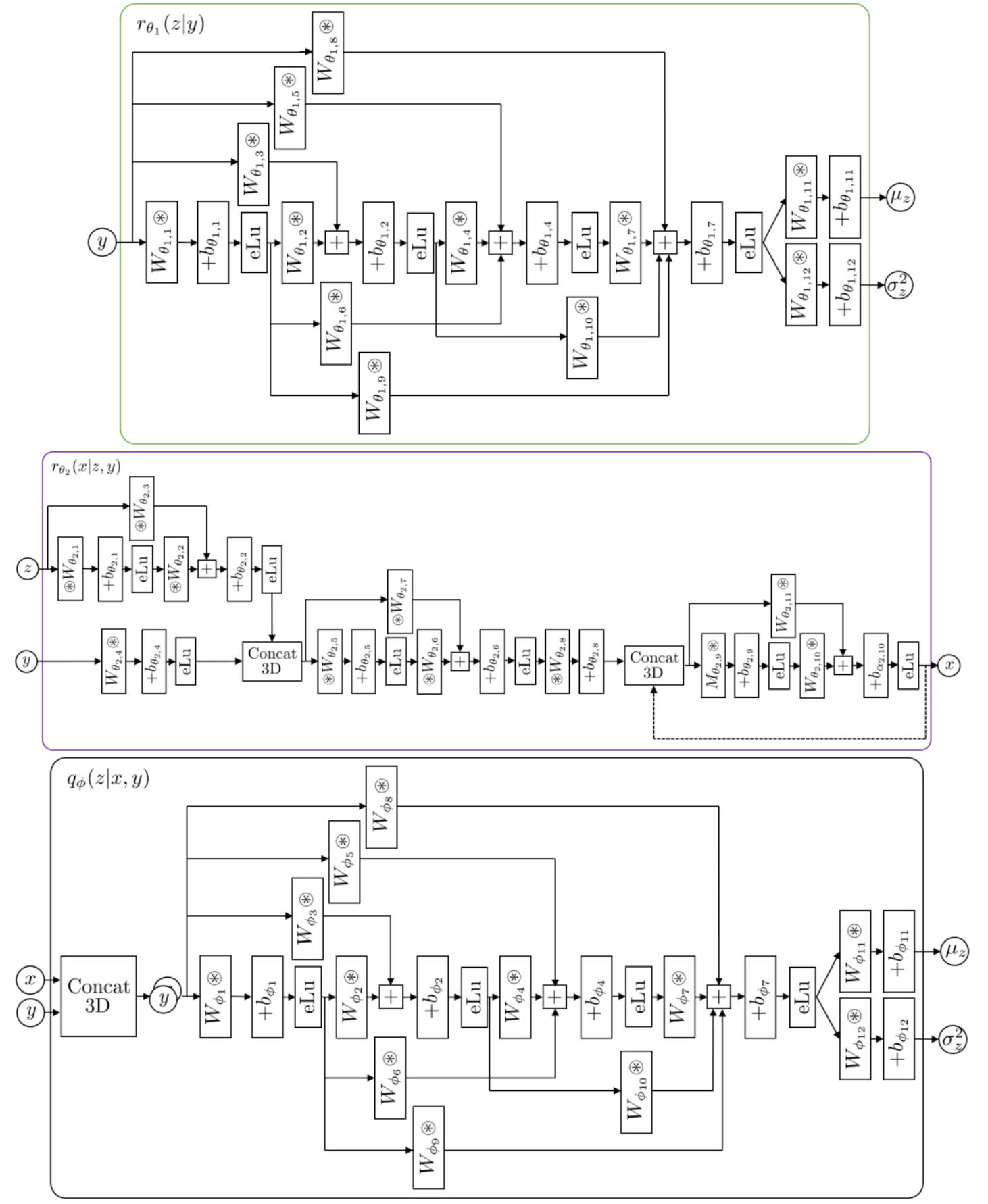}
\caption{Parametric distributions' structures for the convolutional version of the multi-fidelity forward model. $W \circledast$ indicates a convolution with filter bank $W$, while $\circledast W$ indicates a transpose convolution. $M$ indicates the masked convolution part of the PixelVAE model (see \cite{PIXVAE}).}
\label{fig:supp_inverse_conv}
\end{figure*}

Like the multi-fidelity forward model, the inverse model includes three parametric distributions (see figure \ref{fig:inverse}); $p_{\theta_1}(z|y)$,  $p_{\theta_2}(x|y,z)$ and $q_{\phi}(z|x,y)$. As before, two versions of the inverse model model were implemented. In the first, the parametric distributions consist of fully connected layers mapping inputs to outputs' Gaussian moments, from which samples are drawn upon training and inference. These structures are schematically represented in figure \ref{fig:supp_inverse_fullyconn}. In the second, $p_{\theta_1}(z|y)$ and $q_{\phi}(z|x,y)$ consist of deeper convolutional recurrent layers, again mapping mapping inputs to outputs' Gaussian moments, from which samples are drawn upon training and inference. $p_{\theta_2}(x|y,z)$ is similarly built with convolutional layers, but the generation of the final images is performed conditioning on previously predicted adjacent pixels with a masked convolution as described in \cite{PIXVAE}. These structures are instead shown in figure \ref{fig:supp_inverse_conv}.

\section{Details of Experiments}

\subsection{Simulated Experiments}\label{exp_1}

\subsubsection{Qualitative Comparison with Standard CVAE}

Variational models are trained to reconstruct images of faces from blurred and noisy observations. First, CVAEs are trained directly, using $K$ available images and observations as training targets and conditions respectively. Second, the same CVAE models are trained with the proposed framework, making use of the same $K$ paired examples, but including the whole training set of $L = 100,000$ unobserved targets from the CelebA data set and the inaccurate observation model as described in section \ref{sec:technical}.

The first set of experiments shown in figure \ref{fig:inc_fig} was carried out with a $64 \times 64$ down-sampled and centered version of the CelebA data set. Three Gaussian blurring conditions were tested, with increasing PSF width and noise standard deviation. In each case, the PSF and noise where chosen differently for the true transformation, applied to the small set of paired examples and the test data, and an inaccurate observation model, used instead as the low-fidelity model from domain expertise. In the first experiment, the true blurring Gaussian PSF was set to have standard deviation $\sigma_{PSF} = 2px$ and signal to noise ratio (SNR) of $25 dB$, while the low-fidelity model was given $\sigma_{PSF} = 1.5px$ and $SNR = 28dB$. In the second experiment, the true blurring Gaussian PSF was set to have standard deviation $\sigma_{PSF} = 4px$ and signal to noise ratio (SNR) of $16 dB$, while the low-fidelity model was given $\sigma_{PSF} = 3px$ and $SNR = 20dB$. In the third experiment, the true blurring Gaussian PSF was set to have standard deviation $\sigma_{PSF} = 6px$ and signal to noise ratio (SNR) of $8 dB$, while the low-fidelity model was given $\sigma_{PSF} = 4px$ and $SNR = 12dB$.

The multi-fidelity forward model used in these experiment is the convolutional version, the components of which are illustrated in figure \ref{fig:supp_forward_conv}. The inverse model, inferring reconstructed images from blurred observations, is also the convolutional version shown in figure \ref{fig:supp_inverse_conv}, both for the proposed training method and for the CVAE standard training. The sizes of the filter banks $W$ used are reported in table \ref{tab:exp1}.

\begin{center}
\begin{table}
\caption{Filter banks of the multi-fidelity forward model and variational inverse model used in the experiments of figure \ref{fig:inc_fig}. The table reports the filter bank name used in the architectures shown in figures \ref{fig:supp_forward_conv} and \ref{fig:supp_inverse_conv}, filters height$\times$width$\times$number of channels and strides of the convolutions.}
\label{tab:exp1}
\vspace{0.3cm}
  \begin{tabular}{| l | p{2.4cm} | p{2.4cm} || l | p{2.4cm} | p{2.4cm}|}
    \hline
    \textbf{Filters} & \textbf{h $\times$ w $\times$ c} & \textbf{Strides} & \textbf{Filters} & \textbf{h $\times$ w $\times$ c} & \textbf{Strides} \\
    \hline
    $W_{\alpha_{1,1}}$ & $12 \times 12 \times 10$ & $2 \times 2$ & $W_{\alpha_{1,2}}$ & $12 \times 12 \times 10$ & $1 \times 1$ \\
    \hline
    $W_{\alpha_{1,3}}$ & $12 \times 12 \times 10$ & $2 \times 2$ & $W_{\alpha_{1,4}}$ & $12 \times 12 \times 10$ & $2 \times 2$ \\
    \hline
    $W_{\alpha_{1,5}}$ & $12 \times 12 \times 10$ & $4 \times 4$ & $W_{\alpha_{1,6}}$ & $12 \times 12 \times 10$ & $2 \times 2$ \\
    \hline
    $W_{\alpha_{1,7}}$ & $12 \times 12 \times 10$ & $2 \times 2$ & $W_{\alpha_{1,8}}$ & $12 \times 12 \times 10$ & $8 \times 8$ \\
    \hline
    $W_{\alpha_{1,9}}$ & $12 \times 12 \times 10$ & $2 \times 2$ & $W_{\alpha_{1,10}}$ & $12 \times 12 \times 10$ & $4 \times 4$ \\
    \hline
    $W_{\alpha_{1,11}}$ & $12 \times 12 \times 3$ & $1 \times 1$ & $W_{\alpha_{1,12}}$ & $12 \times 12 \times 3$ & $1 \times 1$ \\
    \hline
    $W_{\alpha_{2,1}}$ & $12 \times 12 \times 10$ & $2 \times 2$ & $W_{\alpha_{2,2}}$ & $12 \times 12 \times 10$ & $2 \times 2$ \\
    \hline
    $W_{\alpha_{2,3}}$ & $12 \times 12 \times 10$ & $4 \times 4$ & $W_{\alpha_{2,4}}$ & $12 \times 12 \times 10$ & $2 \times 2$ \\
    \hline
    $W_{\alpha_{2,5}}$ & $12 \times 12 \times 10$ & $1 \times 1$ & $W_{\alpha_{2,6}}$ & $12 \times 12 \times 10$ & $2 \times 2$ \\
    \hline
    $W_{\alpha_{2,7}}$ & $12 \times 12 \times 10$ & $2 \times 2$ & $W_{\alpha_{2,8}}$ & $12 \times 12 \times 3$ & $1 \times 1$ \\
    \hline
    $W_{\alpha_{2,9}}$ & $12 \times 12 \times 3$ & $1 \times 1$ &  &  &  \\
    \hline
    $W_{\beta_{1}}$ & $12 \times 12 \times 10$ & $2 \times 2$ & $W_{\beta_{2}}$ & $12 \times 12 \times 10$ & $1 \times 1$ \\
    \hline
    $W_{\beta_{3}}$ & $12 \times 12 \times 10$ & $2 \times 2$ & $W_{\beta_{4}}$ & $12 \times 12 \times 10$ & $2 \times 2$ \\
    \hline
    $W_{\beta_{5}}$ & $12 \times 12 \times 10$ & $4 \times 4$ & $W_{\beta_{6}}$ & $12 \times 12 \times 10$ & $2 \times 2$ \\
    \hline
    $W_{\beta_{7}}$ & $12 \times 12 \times 10$ & $2 \times 2$ & $W_{\beta_{8}}$ & $12 \times 12 \times 10$ & $8 \times 8$ \\
    \hline
    $W_{\beta_{9}}$ & $12 \times 12 \times 10$ & $4 \times 4$ & $W_{\beta_{10}}$ & $12 \times 12 \times 10$ & $2 \times 2$ \\
    \hline
    $W_{\beta_{11}}$ & $12 \times 12 \times 3$ & $1 \times 1$ & $W_{\beta_{12}}$ & $12 \times 12 \times 3$ & $1 \times 1$ \\
    \hline
    $W_{\theta_{1,1}}$ & $9 \times 9 \times 30$ & $2 \times 2$ & $W_{\theta_{1,2}}$ & $9 \times 9 \times 30$ & $1 \times 1$ \\
    \hline
    $W_{\theta_{1,3}}$ & $9 \times 9 \times 30$ & $2 \times 2$ & $W_{\theta_{1,4}}$ & $9 \times 9 \times 30$ & $2 \times 2$ \\
    \hline
    $W_{\theta_{1,5}}$ & $9 \times 9 \times 30$ & $4 \times 4$ & $W_{\theta_{1,6}}$ & $9 \times 9 \times 30$ & $2 \times 2$ \\
    \hline
    $W_{\theta_{1,7}}$ & $9 \times 9 \times 30$ & $2 \times 2$ & $W_{\theta_{1,8}}$ & $9 \times 9 \times 30$ & $8 \times 8$ \\
    \hline
    $W_{\theta_{1,9}}$ & $9 \times 9 \times 30$ & $4 \times 4$ & $W_{\theta_{1,10}}$ & $9 \times 9 \times 30$ & $2 \times 2$ \\
    \hline
    $W_{\theta_{1,11}}$ & $9 \times 9 \times 3$ & $1 \times 1$ & $W_{\theta_{1,12}}$ & $9 \times 9 \times 3$ & $1 \times 1$ \\
    \hline
    $W_{\theta_{2,1}}$ & $9 \times 9 \times 30$ & $2 \times 2$ & $W_{\theta_{2,2}}$ & $9 \times 9 \times 30$ & $2 \times 2$ \\
    \hline
    $W_{\theta_{2,3}}$ & $9 \times 9 \times 30$ & $4 \times 4$ & $W_{\theta_{2,4}}$ & $9 \times 9 \times 30$ & $2 \times 2$ \\
    \hline
    $W_{\theta_{2,5}}$ & $9 \times 9 \times 30$ & $1 \times 1$ & $W_{\theta_{2,6}}$ & $9 \times 9 \times 30$ & $2 \times 2$ \\
    \hline
    $W_{\theta_{2,7}}$ & $9 \times 9 \times 30$ & $2 \times 2$ & $W_{\theta_{2,8}}$ & $9 \times 9 \times 3$ & $1 \times 1$ \\
    \hline
    $M_{\theta_{2,9}}$ & $9 \times 9 \times 10$ & $1 \times 1$ &  $W_{\theta_{2,10}}$ & $9 \times 9 \times 3$ & $1 \times 1$  \\
    \hline
    $W_{\phi_{1}}$ & $9 \times 9 \times 30$ & $2 \times 2$ & $W_{\phi_{2}}$ & $9 \times 9 \times 30$ & $1 \times 1$ \\
    \hline
    $W_{\phi_{3}}$ & $9 \times 9 \times 30$ & $2 \times 2$ & $W_{\phi_{4}}$ & $9 \times 9 \times 30$ & $2 \times 2$ \\
    \hline
    $W_{\phi_{5}}$ & $9 \times 9 \times 30$ & $4 \times 4$ & $W_{\phi_{6}}$ & $9 \times 9 \times 30$ & $2 \times 2$ \\
    \hline
    $W_{\phi_{7}}$ & $9 \times 9 \times 30$ & $2 \times 2$ & $W_{\phi_{8}}$ & $9 \times 9 \times 30$ & $8 \times 8$ \\
    \hline
    $W_{\phi_{9}}$ & $9 \times 9 \times 30$ & $4 \times 4$ & $W_{\phi_{10}}$ & $9 \times 9 \times 30$ & $2 \times 2$ \\
    \hline
    $W_{\phi_{11}}$ & $9 \times 9 \times 3$ & $1 \times 1$ & $W_{\phi_{12}}$ & $9 \times 9 \times 3$ & $1 \times 1$ \\
    \hline
  \end{tabular}
  \vspace{-0.3cm}
  \end{table}
\end{center}

\subsubsection{PSNR and ELBO versus Number of Examples}

The set of experiments giving results shown in figure \ref{fig:Kvar} is carried out on a $32 \times 32$ down-sampled version of the CelebA data set. Images are blurred with a Gaussian PSF having a standard deviation of $2$ pixels. As before, the standard CVAEs are trained with the $K$ image-observation pairs alone. The proposed framework is then applied in each condition, exploiting the same $K$ paired images and observations, $L = 100,000$ unobserved target examples and an inaccurate observation model. Two different inaccurate observation models are used; a more accurate one with $10\%$ under-estimation of PSF width and noise level and a less accurate one, having $40\%$ under-estimation. After training each model, reconstructions are performed with $2,000$ test examples and two quantitative measures are extracted: (i) the average peak signal to noise ration (PSNR) between the pseudo-maximum reconstructions and the original images and (ii) the evidence lower bound (ELBO). The former serves as a measure of deterministic performance, giving an index of similarity between the ground truth and the most likely reconstruction. The latter is a measure of probabilistic performance, as it is an approximation to the likelihood assigned by the model to the ground truths and consequentially is an index of how well the distribution of solutions to the inverse model is captured.

The forward multi-fidelity model used in the proposed methods is built with the simple fully connected structures of figure \ref{fig:supp_forward_fullyconn}. The comparison CVAE and the inverse models are identical and are built with the fully connected structures of figure \ref{fig:supp_inverse_fullyconn}. Multi-fidelity forward models were built to have $300$ hidden units in all deterministic layers, while the latent variable $w$ was chosen to be $100$-dimensional. The inverse models, both for the proposed framework and the comparative CVAE, were built with $2500$ hidden units in the deterministic layers and latent variables $z$ of $800$ dimensions.

\subsubsection{Qualitative Comparison with Alternative Training Strategies}

The results of figure \ref{fig:diff} were obtained by reconstructing from blurred $64 \times 64$ CelebA images, blurred with a Gaussian PSF having standard deviation of $4$ pixels and additive gaussian noise, corresponding to a SNR of $16dB$. The inaccurate observation model was instead given a PSF with standard deviation of $3$ pixels and no additive noise.

The multi-fidelity forward model used in these experiments was of the convolutional form shown in figure \ref{fig:supp_forward_conv}, with the filter structures reported in table \ref{tab:exp1}.

\subsubsection{Quantitative Comparison with Alternative Training Strategies}

Four different degradation conditions where tested, applying the following four degradations to $32 \times 32$ CelebA and CIFAR10 images:

\begin{itemize}
    \item \textbf{$\times$ 2 Down-Sampling.} The true transformation applied to the $K$ observed images consists of a $\times 2$ down-sampling of the images in each dimension and a subsequent blurring with a PSF having standard deviation $1.4$ pixels. The low-fidelity accessible model down-samples by $2$, but does not apply any blurring afterwards (i.e. the source of inaccuracy in the known forward model derives from ignoring blurring).
    \item \textbf{Partial Occlusion.} In the true transformation applied to the $K$ observed images, a rectangular section of $8 \times 11$ pixels is set to zero in a given position in all images. The low-fidelity model places instead a $5 \times 15$ at random with a difference in central position of $dy =2$ and $dx = -2$.
    \item \textbf{Gaussian Blurring, $\sigma_{PSF} = 2.5 px$.} The true transformation blurs the images with a PSF having standard deviation $\sigma_{PSF} = 2.5$ pixels and additive Gaussian noise at $12dB$. The low-fidelity analytical model instead blurs the images with a PSF having standard deviation $\sigma_{PSF} = 1.5$ pixels and does not add any noise.
    \item \textbf{Gaussian Blurring, $\sigma_{PSF} = 1.5 px$.} The true transformation blurs the images with a PSF having standard deviation $\sigma_{PSF} = 1.5$ pixels and additive Gaussian noise at $16dB$. The low-fidelity analytical model instead blurs the images with a PSF having standard deviation $\sigma_{PSF} = 1$ pixels and does not add any noise.
\end{itemize}

The forward multi-fidelity model used in the proposed methods is built with the simple fully connected structures of figure \ref{fig:supp_forward_fullyconn}. All inversion models, competitive and proposed, were implemented with the fully connected version of the inverse model given in figure \ref{fig:supp_inverse_fullyconn}. Multi-fidelity forward models were built to have $300$ hidden units in all deterministic layers, while the latent variable $w$ was chosen to be $100$-dimensional. The inverse models, both for the proposed framework and the comparative training methods, were built with $2500$ hidden units in the deterministic layers and latent variables $z$ of $800$ dimensions.

\subsection{Holographic Image Reconstruction}\label{exp_2}

We provide here more details on the HIO algorithm. The HIO algorithm is a Fourier transform-based method for holographic reconstruction where some constraints are used as support. In our case we have access to the amplitude at the camera plane and we assume that the phase at the DMD plane is uniform accross all micromirrors. The HIO algorithm starts with a random guess of the phase of the recorded image at the camera, performs an inverse Fourier transform to obtain a guess of both amplitude and phase at the DMD plane, and replaces the obtained phase with a uniform phase (one of our constraints). At this point, further constraints are added e.g. there is only image information at the central $N \times M$ pixels of the image (with $N,M$ being arbitrary). After that, a forward Fourier transform is performed and the corresponding amplitude is replaced by the image recorded by the camera. This process is repeated iteratively. The problem is that if the recorded image is saturated and down-sampled, the iterative process breaks after the first iteration. As a consequence, it is impossible for the algorithm to converge towards a solution close to the ground truth. This is precisely what we observe in Figure~\ref{fig:FFT_main}(c), where the HIO algorithm simply predicts spots at some positions.

\subsection{ToF Diffuse Imaging}\label{exp_3}

The comparative iterative method was taken from \cite{DOT_NAT}, reproducing exactly the main results therein. For the proposed variational method, only the first $15$ frames of the recorded experimental video were used as observation, as most of the information is contained in the rising front of the signal and around the peak. Consequentially, the corresponding frames in the two simulations (high and low fidelity) were used to train the model. The forward multi-fidelity model for the proposed variational method was built with the fully connected structures shown in figure \ref{fig:supp_forward_fullyconn}, with all deterministic intermediate layers having $3000$ hidden units and latent variables $w$ having $100$ dimensions. The inverse model was also constructed using fully connected structures, as shown in figure \ref{fig:supp_inverse_fullyconn}, with all deterministic intermediate layers having $1500$ hidden units and latent variables $z$ having $800$ dimensions.

\newpage

\bibliography{sample}

\end{document}